\documentclass[sigconf]{acmart}

\AtBeginDocument{%
  }

\setcopyright{rightsretained}

\acmYear{2023}\copyrightyear{2023}
\acmConference[MMSys '23]{Proceedings of the 14th ACM Multimedia Systems Conference}{June 7--10, 2023}{Vancouver, BC, Canada}
\acmBooktitle{Proceedings of the 14th ACM Multimedia Systems Conference (MMSys '23), June 7--10, 2023, Vancouver, BC, Canada}
\acmPrice{15.00}
\acmDOI{10.1145/3587819.3590973}
\acmISBN{979-8-4007-0148-1/23/06}

\acmSubmissionID{67}

\usepackage{times}
\usepackage{epsfig}
\usepackage{graphicx}
\usepackage{amsmath}
\usepackage{booktabs}

\usepackage{subfigure}
\usepackage{colortbl}
\usepackage{bbm}
\usepackage{units}
\usepackage{pifont}
\usepackage{xspace}
\usepackage{tablefootnote}
\newcommand{\cmark}{\ding{51}}%
\begin{document}

\title{Color-aware Deep Temporal Backdrop Duplex Matting System}

\author{Hendrik Hachmann and Bodo Rosenhahn}
\affiliation{%
  \institution{Institute for Information Processing (tnt) /
L3S - Leibniz University Hannover }
  \streetaddress{Appelstr. 9a}
   \city{Hannover}
   \country{Germany}}
 \email{{hachmann, rosenhahn}@tnt.uni-hannover.de}

\renewcommand{\shortauthors}{Hendrik Hachmann and Bodo Rosenhahn}

\makeatletter

\DeclareRobustCommand\onedot{\futurelet\@let@token\@onedot}
\def\@onedot{\ifx\@let@token.\else.\null\fi\xspace}

\def\eg{\emph{e.g}\onedot} \def\Eg{\emph{E.g}\onedot}
\def\ie{\emph{i.e}\onedot} \def\Ie{\emph{I.e}\onedot}
\def\cf{\emph{c.f}\onedot} \def\Cf{\emph{C.f}\onedot}
\def\etc{\emph{etc}\onedot} \def\vs{\emph{vs}\onedot}
\def\wrt{w.r.t\onedot} \def\dof{d.o.f\onedot}
\def\etal{\emph{et al}\onedot}
\makeatother

\begin{abstract}
Deep learning-based alpha matting showed tremendous improvements in recent years, yet, feature film production studios still rely on classical chroma keying including costly post-production steps. This perceived discrepancy can be explained by some missing links necessary for production which are currently not adequately addressed in the alpha matting community, in particular foreground color estimation or color spill compensation. We propose a neural network-based temporal multi-backdrop production system that combines beneficial features from chroma keying and alpha matting. Given two consecutive frames with different background colors, our one-encoder-dual-decoder network predicts foreground colors and alpha values using a patch-based overlap-blend approach. The system is able to handle imprecise backdrops, dynamic cameras, and dynamic foregrounds and has no restrictions on foreground colors. We compare our method to state-of-the-art algorithms using benchmark datasets and a video sequence captured by a demonstrator setup. We verify that a dual backdrop input is superior to the usually applied trimap-based approach. In addition, the proposed studio set is actor friendly, and produces high-quality, temporal consistent alpha and color estimations that include a superior color spill compensation.
\end{abstract}



\maketitle

\newcommand{\mywidtht}{0.18}
\newcommand{\myheightt}{0.18}
\begin{figure}[t]
  \begin{center}
  \begin{subfigure}[Foreground]
    {
    \centering
    \includegraphics[height=\myheightt\textheight]{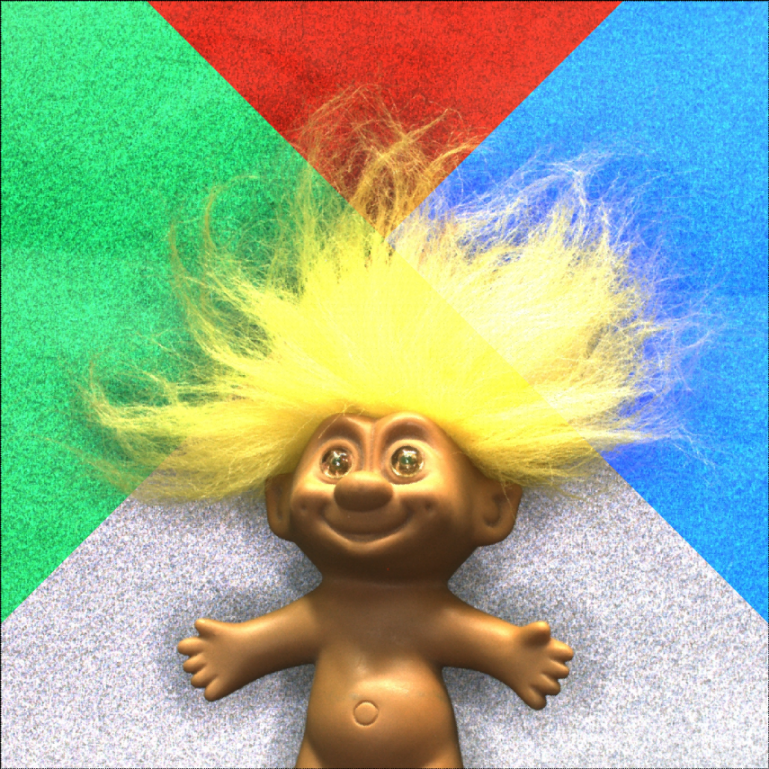}
    }	    
  \end{subfigure}
  \begin{subfigure}[RGBa and trimap]
    {
    \centering
    \includegraphics[height=\myheightt\textheight]{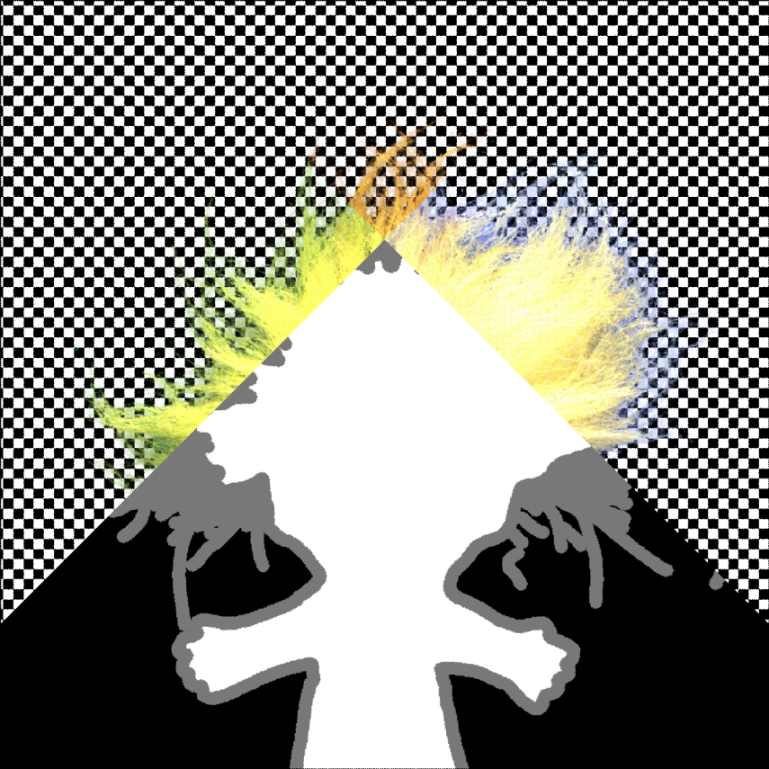}
    }
  \end{subfigure}
  \end{center}
  \caption{(a) a collage of 4 images of a troll in front of an retroreflective background, that has been illuminated with different colors. ((b), upper area) three compositions of an alpha-blended troll in front of a checkerboard background. The colored fringes are a problem called \textit{color spill}, which is background color incorrectly extracted as foreground. ((b), lower area) a so-called \textit{trimap}, that signals foreground, background and a gray area, the latter may contain transparencies.\vspace{-1mm}}
  \label{colorspill}
\end{figure}
\section{Introduction}
\label{sec:intro}
During the production of the 2019 Disney+ series ``The Mandalorian'', Industrial Light \& Magic introduced StageCraft \cite{Stagecraft}, a very high-definition LED video wall, in which visual effects are displayed on the wall, directly captured by the camera and thus appear in the footage. In this setup, traditional green screens are rendered unnecessary and no color keying or \textit{pulling the matte} of objects or actors is needed, which simplifies the production and reduces visual effects (VFX) shot costs. Some say this may leave green screen technology obsolete. However, the option to change VFX at post-production is hereby abandoned. Nevertheless, the integration of similar video walls in studios may create interesting options for improved matting applications, one of which is proposed in this paper.

Foreground transparency estimation is called \textit{alpha matting}, with $\alpha$ being the amount of transparency for each pixel. $\alpha = 0$ denotes fully transparent foregrounds and $\alpha = 1$ meaning opacity. Formally, the matting equation
\begin{equation}
 C = \alpha C_{fg} + (1-\alpha) C_{bg}
 \label{mattingequation}
\end{equation}
needs to be solved, which is an ill-posed problem. The image $C$ is often named \textit{composition}, since it is an $\alpha$-weighted superposition of the foreground color $C_{fg}$ and the background color $C_{bg}$. Chroma keying refers to blue or green screen matting frequently applied in feature film production, a technique in which the color of the background is a priori given and thus key to the matting task. In production, the foreground color $C_{fg}$ needs to be estimated as well as $\alpha$, resulting in an RGBa stack.

A large number of green screen applications exist. They are used e.g.\ in news studios, weather forecasting, and film production. Chroma keying works well in expensive studios with highly controlled illumination. However, it poses algorithmic limitations as well as undesirable interference with actors.

There is obviously the limitation that the background color is to be avoided in the foreground. Furthermore, as can be seen in Figure \ref{colorspill}, there is a problem called \textit{color spill}: Visible background colored fringes shine through the foreground object at transparent points or are projected onto the foreground object from the side. Color spill is very noticeable in the composition with an alternative background and should be avoided or compensated for as much as possible. 

\begin{figure}[t]
  \begin{center}
    \includegraphics[width=\linewidth]{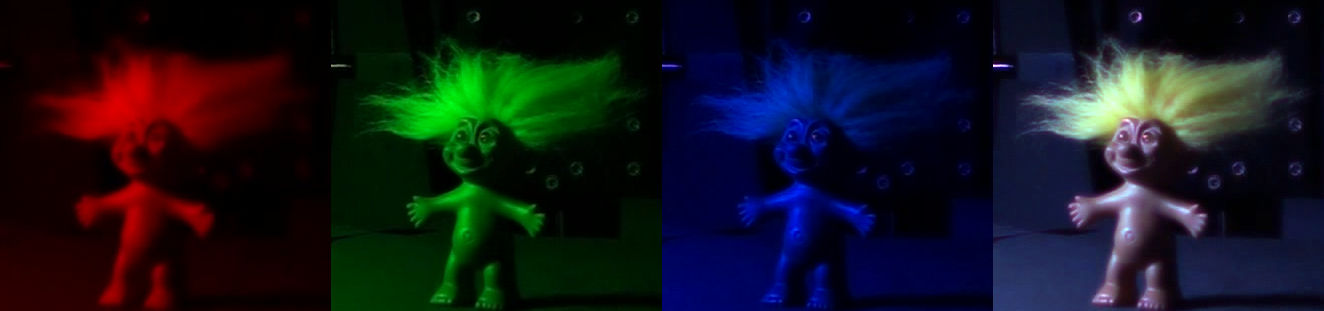}
  \end{center}
  \caption{Images of a troll with yellow hair, that are captured with the same, static camera under varying monochromatic illuminations. It can be seen that the troll under green illumination appears detailed, while the red and blue illuminated images are blurry. This effect is called chromatic aberration, an effect caused by dispersion at the camera lens. The superposition of the three mono chromatic images is shown at the right. Here, the effect of chromatic aberration is visible as purple fringes.\vspace{-1mm}}
  \label{aberration1}
\end{figure}

\begin{figure}[t]
  \begin{center}
    \includegraphics[width=\linewidth]{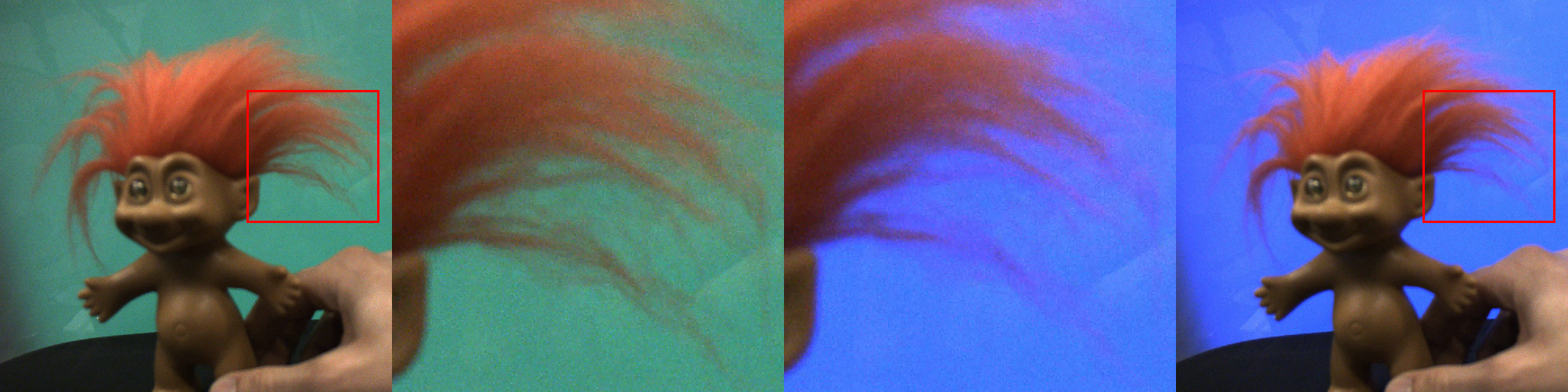}
  \end{center}
  \caption{Effect of chromatic aberration illustrated by two consecutive frames taken form a troll sequence. The red hair is better visible in front of the green background compared to the blue background. Consequently, the hair strands on the left are perceived less transparent by humans and estimated by algorithms, which results in flickering alpha predictions in an alternating background color system.\vspace{-1mm}}
  \label{aberration2}
\end{figure}

In addition, matting is challenging if chromatic aberration \cite{Korneliussen:2014} occurs. While this effect is often neglected in alpha matting literature, it is increasingly important if camera resolution increases or if consumer cameras are used. The effect of chromatic aberration is caused by dispersion at the camera lens. It can be seen in Figure \ref{aberration1}, in which images of a troll are captured with monochromatic illumination. It can be observed that under green illumination the troll appears detailed, while the image becomes blurred or out of focus with red and blue illumination. This effect leads to colored fringes also known as rainbow edge in images with white illumination. In context of chroma keying systems using green or blue screens this effect can lead to a misperception as can be seen in Figure \ref{aberration2}. The red hair of the troll is perceived less transparent in front of a green background compared to a blue background. 

A \text{color-aware} matting system needs to predict foreground colors along with alpha mattes, while compensating color spill and being robust against chromatic aberration. 

We summarise another category of problems as human discomfort impacting involved persons, i.e.\  newscasters or actors. Green screens must be illuminated very homogeneously. This is often achieved by strong illumination, which causes huge amounts of spotlights to heat up the room. In addition, humans feel the artificial environment, renders them disoriented and it is difficult for actors to put themselves into a scene, lacking so-called \textit{immersive feedback}. That is why i.e.\ markers are often used so that actors at least look in the right direction of the VFX content. Markers, on the other hand, have to be masked to not influence chroma keying. 

In this paper, we propose a temporally alternating backdrop matting system permitting dynamic cameras and foregrounds, alleviating foreground color restrictions, and allowing imprecise backings. The system deploys a one-encoder-dual-decoder neural network, that in an overlap-blend approach produces high-quality alpha and color estimation, including an advanced color spill compensation. The resulting simplification of studio sets along with high-quality matting can reduce production and post-processing costs. In addition, our system provides an actor-friendly environment with visual clues without any color restriction, enabling the actor to dive into the scene while performing.  

The contributions of this paper are summarised by:
\begin{itemize}
    \item We present a fully-functional temporal backdrop duplex setup, consisting of a camera and FPGA controlled LED panels synchronized @\unit[100]{fps}.
    \item Our hardware setup demonstrates the feasibility of an actor friendly studio set.
    \item A novel one-encoder-dual-decoder neural network architecture allows prediction of RGBa foregrounds from two consecutive frames with alternating backdrop color.
    \item The network handles dynamic scenes by combination of an inner patch prediction and an overlap-blend subdivision.
    \item We quantify the benefit of using dual backdrops instead of trimaps as input for neural alpha-matting.    
    \item An automatic advanced color spill suppression method is proposed for post-production. 
\end{itemize}

In the remainder of this paper, we review related work, describe our method and compare the performance to state-of-the-art approaches.
\section{Related work}
\label{sec:SOTA}
\textbf{Studio sets:} While the green or blue color is key for chroma keying, artificial homogeneous walls often leave actors without orientation. To regain orientation markers can be inserted that are later, sometimes even manually, masked out and removed from the footage. Tzidon and Tzidon \cite{Tzidon:1999} introduce a synchronized time duplex system in which markers are projected to a green screen at the readout time of the camera, generally called \textit{blanking time}. Consequently, those markers do not appear in the footage at all. Vidal and Lafuente \cite{Vidal:2016} use video projectors to add amplitudes of green to a green screen, without impacting the chroma keying. Within certain limits, these shades of green can give monochromatic visual clues to actors. Furthermore, Borja Vidal \cite{Vidal:2012} uses polarized light and polarization filters to provide immersive feedback in combination with retroreflective screens. In this studio, the content is projected onto the background but is filtered so that the camera does not capture it. Grau \etal \cite{Grau:2004} use additional cameras to locate and track the actors' heads, in order to render VFX content based on the location of the actor. In this system, the actor sees view-dependent VFX content without geometric distortions to increase immersiveness into the scene. 

\textbf{ Matting methods:} Matting is a long-studied research field. In 2007, Wang and Cohen \cite{Wang:2007} published a survey paper on image and video matting, in which they describe and compare 50 matting approaches. 
The most common scenario is to predict the alpha matte given an input image and a trimap (see Figure \ref{colorspill}), in which pure foreground and pure background are marked as well as a gray area, which is an unknown region that may contain transparencies. Trimaps are often seen as user input. For this matting task, there is an online benchmark by Rhemann\footnote{\url{alphamatting.com}} \etal \cite{Rhemann:2009}, that is currently comparing 68 algorithms. 

To our knowledge, the first deep learning-based matting algorithm is \textit{deep image matting} by Xu \etal \cite{Xu:2017}. They manually create a dataset with ground truth alpha mattes and train a fully convolutional neural network, that given a stack of RGB images and a trimap predicts an alpha matte. The network consists of an encoder-decoder stage predicting a coarse alpha matte and a refinement stage that locally improves the results. Since then researchers argue that the natural structure of foregrounds is inherently learned by neural networks which provides superior performance compared to traditional alpha matting. Recent approaches such as Sun \etal \cite{Sun:2021} combine the matting with a classification task. This semantic image matting uses multiple object categories and individual matting networks are trained. Given hardware memory limitations matting on high-resolution images becomes challenging, which is why Yu \etal \cite{Yu:2020} introduce a patch-based method in which query patches are compared to context patches from different image regions to increase intra-image consistency. Similarly, consistency across scales or hierarchical structures are optimized \cite{Li:2020b,Qiao:2020}. Animal matting is particularly challenging due to fur and camouflage effects. Thus Li \etal \cite{Li:2020} introduce a matting network in which two separate glance and focus networks work together combining the task of recognizing animals and locally extracting fur details.

In feature film production, a trimap is not given. Instead, the color key separates the foreground from the background. A coarse foreground estimation can be generated by different means: Recent progress in face detection initiated a sequence of portrait matting publications \cite{Shen:2016,Li:2021,Ke:2020,Ke:2022}, background subtraction is adopted by Sengupta \etal \cite{Sengupta:2020}, saliency maps by Gupta and Raman \cite{Gupta:2017}, and attention by Zhou \etal \cite{Zhou:2021} and Zhang \etal \cite{Zhang:2021}.

Alpha matting can also be used on videos. Erofeev \etal \cite{Erofeev:2015} maintain a video matting benchmark\footnote{\label{note1}\url{videomatting.com}} created by triangulation in a stop motion fashion. In video matting, typically information is propagated from one frame to the next. For this task, rotoscoping can be applied which is the tracing of shapes or foregrounds in a sequence of images. Agarwala \etal \cite{Agarwala:2004} reduce the manual work of a human in the loop by semi-automatic rotoscoping. Today, this process is automated \cite{Perez-Rua:2020,Backes:2019,Cao:2019} and temporal consistency enforced \cite{Lee:2010, Shahrian:2014,Sun:2021b,Zhang:2021,Lin:2022}.

Many matting approaches predict mattes only. However, for most applications this represents just one of two parts, since the foreground colors need to be estimated as well. Occurring color spill, as a result of imprecise foreground colors is often seen as an independent problem. FBA matting by Forte \etal \cite{Forte:2020}, SIM by Sun \etal \cite{Sun:2021} and the method of Hou and Liu \cite{Hou:2019} are neural networks that simultaneously predict alpha and foreground colors. FBA is currently the leading algorithm on the benchmark of Erofeev \etal \cite{Erofeev:2015}.

The effect of color spill (see Figure \ref{colorspill}) for blue screen keying and a compensation technique was published as early as 1977 by Petro Vlahos \cite{Vlahos:1977}. Since then, the problem has not been fully solved, especially for non-perfect backing information. A common concealment practice is to just reduce the saturation of the foreground color in transparent areas, since a gray color spill attracts less attention. In feature film production color spill removal still requires manual work. More recently, Teng \etal \cite{Teng:2017} introduce a matting method for non-uniform illuminated blue screens.

As a note, convolutional neural networks have successfully been applied to matching and optical flow estimation as in FlowNet \cite{Fischer:2015}, FlowNet 2.0 \cite{Ilg:2017} and deep convolutional matching \cite{Revaud:2015, Thewlis:2016}. All of them can jointly process information that is spatially separated.

According to the survey of Wang and Cohen \cite{Wang:2007} our matting approach would be classified as ``matting with extra information''. Those matting algorithms e.g.\ use flash image pairs as Sun \etal \cite{Sun:2006}, camera arrays as Neel \etal \cite{Neel:2006}, defocussing as McGuire \etal \cite{McGuire:2005} or passive polarization as McGuire \etal \cite{McGuire:2006}. Others use, as we do, multiple backgrounds with changing colors as Smith and Blinn\cite{Smith:1996} and Grundh\"{o}fer \etal \cite{Grundhoefer:2010}. These two methods share a common hardware setup with our approach. Therefore, the next paragraph presents them in detail and they are included in our experimental evaluation (cf.~Section \ref{sec:results}). 

Smith and Blinn \cite{Smith:1996} propose a system directly linked to the matting equation \ref{mattingequation}, which becomes overdetermined and thus solvable for static scenes if two known backgrounds are used. The corresponding method is called \textit{triangulation}. For each pixel it can be implemented as a system of linear equations and is frequently used to generate ground truth datasets. Triangulation can be used to exactly calculate alpha values and foreground colors. The method is also applied by Erofeev \etal \cite{Erofeev:2015} and by Rhemann\footnote{\url{alphamatting.com}} \etal \cite{Rhemann:2009}. 

Another multi-background matting system is proposed by Grundh\"{o}fer \etal \cite{Grundhoefer:2010}, in which they are chroma keying video frames with alternating complementary background colors. In offline mode, trimaps are created by the color difference of backgrounds and Bayesian matting \cite{Chuang:2001} is applied. For static scenes, similar to Smith and Blinn \cite{Smith:1996} this system can create perceptual high-quality mattes by superposition of two mattes since the color spill in both backdrop colors adds up to a neutral ``white''. Being aware that foreground movements introduce errors they apply a seam color compensation heuristic that conceals errors. 

While these temporal backdrop systems are similar in hardware, our deep learning-based matting technique is capable of handling moving foregrounds and backgrounds, is not restricted to precise knowledge of backing colors, and is superior in color spill compensation. Internally our proposed matting system can partly be seen as a frame-wise registration system, followed by matting of registered foregrounds with two known backings which can then be done flawlessly.

\section{Method}
\label{sec:method}

\subsection{Time duplex system}
\begin{figure}[ht]
\begin{center}
 \includegraphics[width=0.98\linewidth]{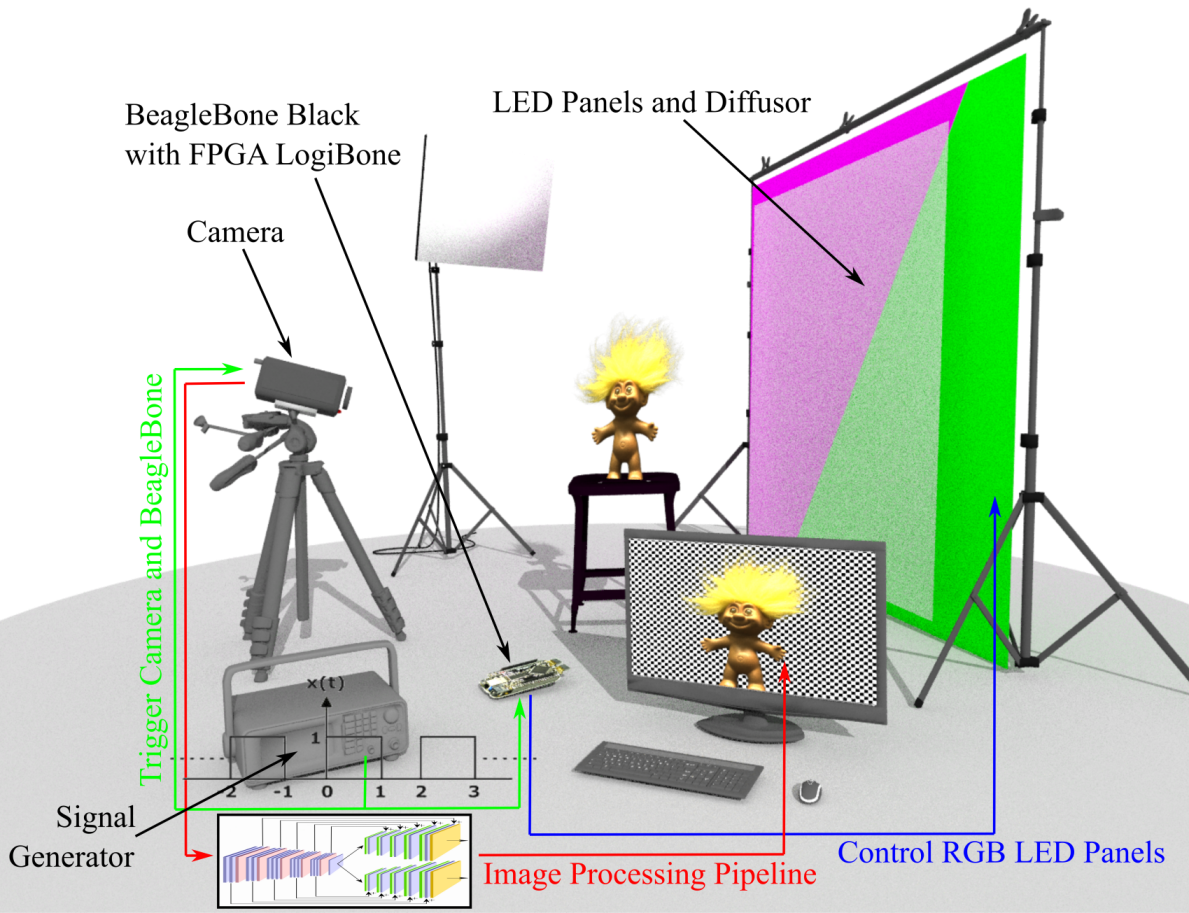}
 \end{center}
 \caption{Studio setup: a camera captures a person or an object in
 front of an LED panel wall with a diffusor, which can provide chroma keying information, e.g.\ a green or blue screen for matting. In time duplex VFX content or markers for actors are displayed on the same wall as well. The panels are controlled by a BeagleBone and the system is synchronized by a signal generator. For each frame, the image processing pipeline creates a composition of the extracted foreground and a new background.\vspace{-1mm}}
 \label{setup}
\end{figure}

As illustrated in Figure \ref{setup}, our proposed studio hardware setup consists of a freely moving global shutter camera (a FLIR ORX-10g-51S5-C set to a resolution of 2448$\times$1600 @\unit[100]{fps}), a diffusor and a synchronized RGB LED wall. The diffuser smoothes individual LED spots of the \unit[6]{mm} pitch modules so that individual LEDs are not recognizable in the camera image. Our demonstrator consists of 4 panels each with 32$\times$32 individually controllable LEDs. The panels are aligned to a roughly \unit[50]{cm} by \unit[50]{cm} ``wall'', which, however, can easily be increased to any desired size. The panels are controlled by a BeagleBone Black and a LogiBone with a Xilinx FPGA according to a description by Glen Akins \cite{Akins:2014}. A signal generator synchronizes the camera and the BeagleBone. The LED wall displays the following sequence: a homogeneous green screen, VFX content, and a homogeneous blue screen followed by VFX content. Then the cycle starts again from the beginning. The system is synchronized so that the camera only captures the green and blue screen during the \unit[1]{ms} exposure time respectively, but not the VFX content, which is shown for \unit[9]{ms} during the camera's blanking time. The human eye, on the other hand, interpolates the full \unit[10]{ms} and therefore mainly perceives the VFX content. Using our demonstrator, we simulate VFX content by a homogeneous red background, which at \unit[100]{fps} is visible to us without any flickering on our wall. In detail, our panel illuminates only four of the 32 LED rows simultaneously, while the other 28 rows are switched off, which is called a 1:8 scan rate. Our FPGA implementation ensures that all rows are turned on for the same amount of time within the \unit[1]{ms} global shutter exposure time, which is why we can achieve a homogeneous illumination across the whole panel. To sum up, our system can provide chroma keying information to the camera as well as different information for actors, e.g. VFX content or markers.
\subsection{Multiple backdrop matting}
\begin{figure}[t]
\begin{center}
 \includegraphics[width=0.99\linewidth]{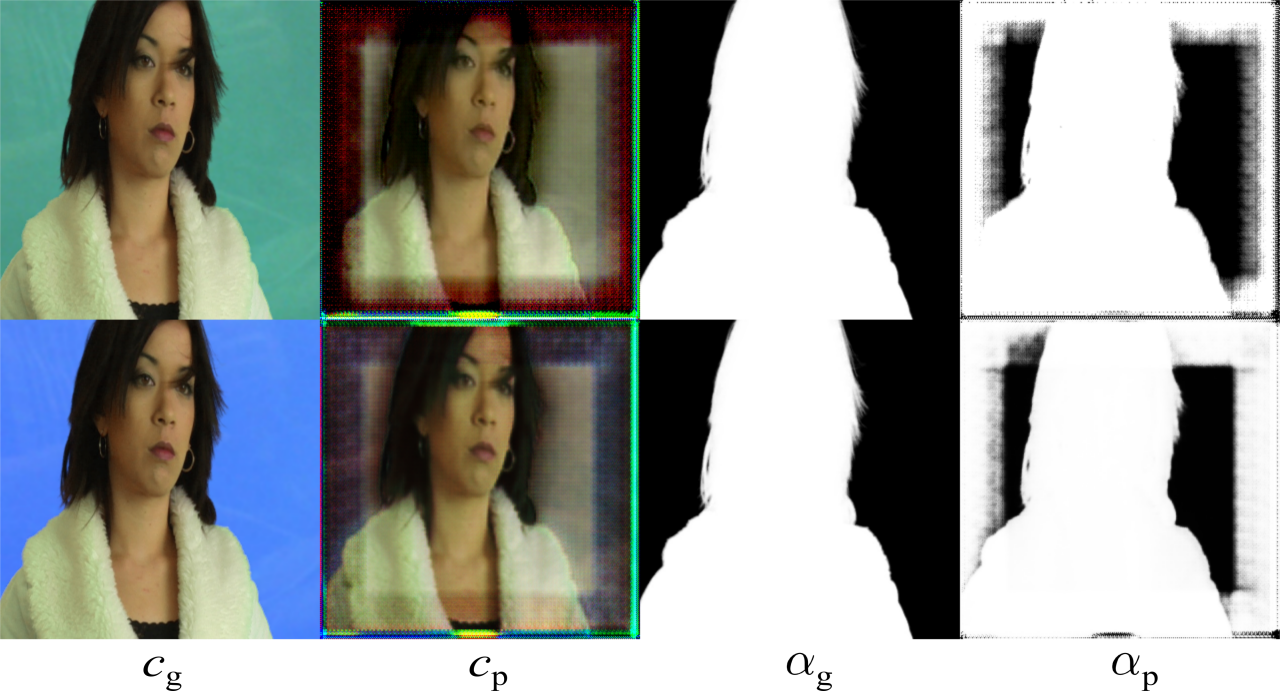}
\end{center}
  \caption{Comparison of foreground color $c_p$ and alpha $\alpha_p$ predictions and the ground truth ($c_g,\alpha_g$). The loss of our neural network is only active at the center of each patch $p$, which we call region $\gamma$, leaving a 50 pixel ``don't care''-area at the boundaries. It is very important to note, that both color predictions $c_p$ (top and bottom) do not contain green or blue color spill in the inner patches.}
  \label{predictions}
\end{figure}

 \textbf{Dataset:} Our image processing pipeline (Figure \ref{setup}) is built on a deep neural network that predicts alpha mattes based on incoming camera frames. This network is trained on the Adobe dataset \cite{Xu:2017}, which provides 432 foreground images with corresponding ground truth alpha mattes. In order to adapt to our use-case, we create our own background dataset, which either includes pure green ($RGB = [0,255,0]$) and pure purple ($RGB = [255,0,255]$) backgrounds or consists of two consecutive frames taken from a 5000 frame video. This video is acquired with our demonstrator that captures the LED panel wall from different angles, resulting in green and blue backings as can be seen in Figure \ref{predictions}, \ref{test_dataset} and \ref{fig:comparison}. These real-world backgrounds have challenging properties: their color varies i.e.\ because of unwanted reflections, changing viewpoints, and noise from the camera's image sensor. These fore- and backgrounds are strictly divided into training and validation samples, with a split of 80\% to 20\%. Random selection of foreground-background combinations creates a 34560 samples training set and a 8640 samples validation set. \\

\begin{figure}[b]
\begin{center}
 \includegraphics[width=0.99\linewidth]{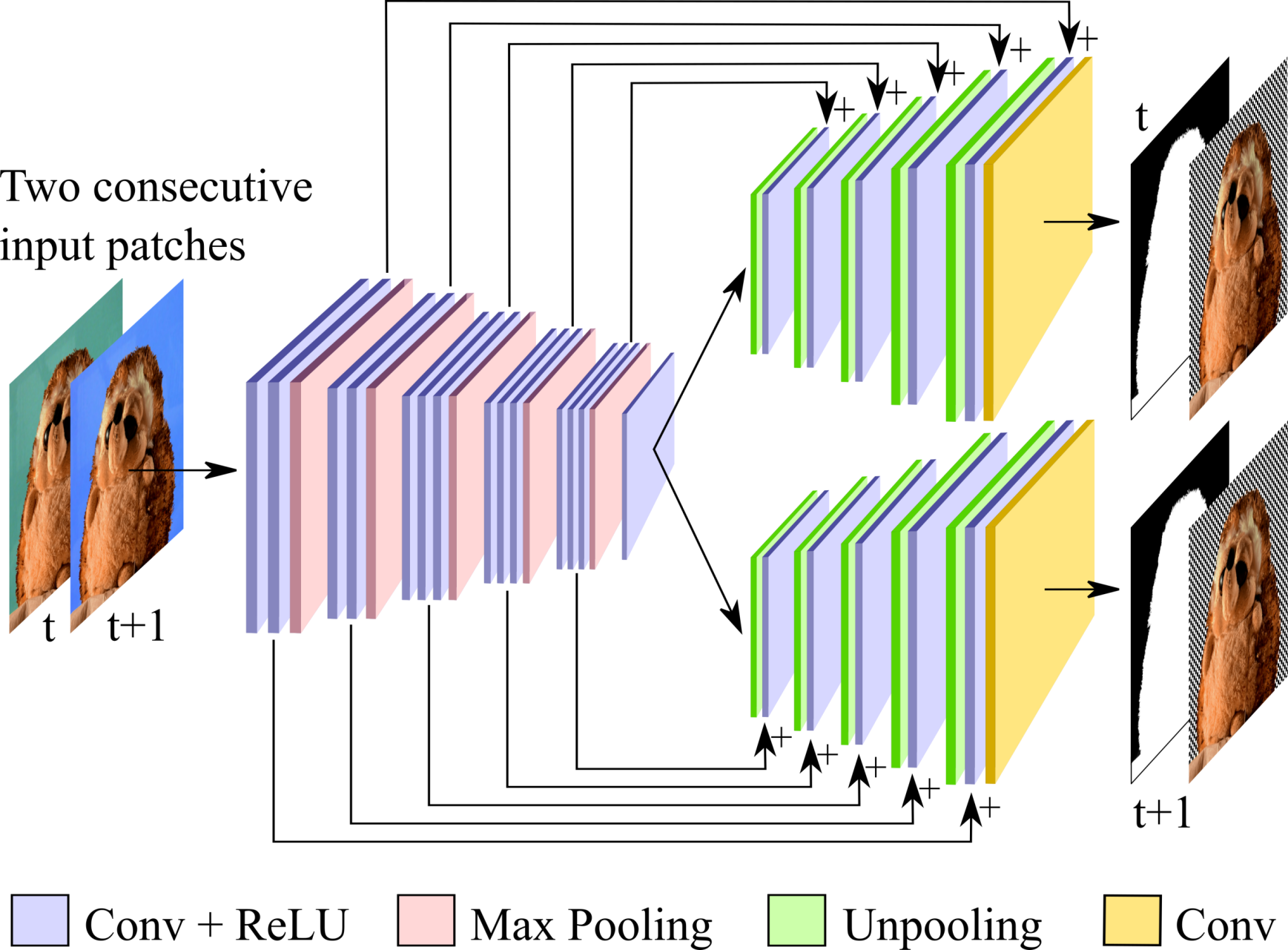}
  \end{center}
 \caption{Given two sequential input patches, our convolutional neural network predicts an alpha matte as well as foreground colors for each frame, using a single-encoder-dual-decoder architecture with skip
 connections.}
 \label{network}
\end{figure}

\textbf{Deep neural network:} The focus of this work is on the overall matting system rather than a highly optimized network architecture. Thus, our network structure (Figure \ref{network}) is by design very traditional and close to Xu \etal's \cite{Xu:2017} encoder-decoder stage. The input to our network are two patches $p_1$ and $p_2$, which are extracted from consecutive RGB frames ($f_1$ and $f_2$) with different background colors. The decoder is realized by two separate branches that each outputs 4-channels consisting of RGB foreground color and an alpha estimation. We add the second decoder branch because our network processes patches from consecutive frames $f_1$ and $f_2$. The encoder and decoder are identical in construction to Xu \etal \cite{Xu:2017}, using 14 convolutional layers with ReLUs and 5 max-pooling layers. Each of the two decoders has 6 convolutional layers with ReLUs and 5 unpooling layers. We add skip connections between feature maps of encoder and decoder. 

\begin{figure}[ht]
 \begin{center}
 \includegraphics[width=0.99\linewidth]{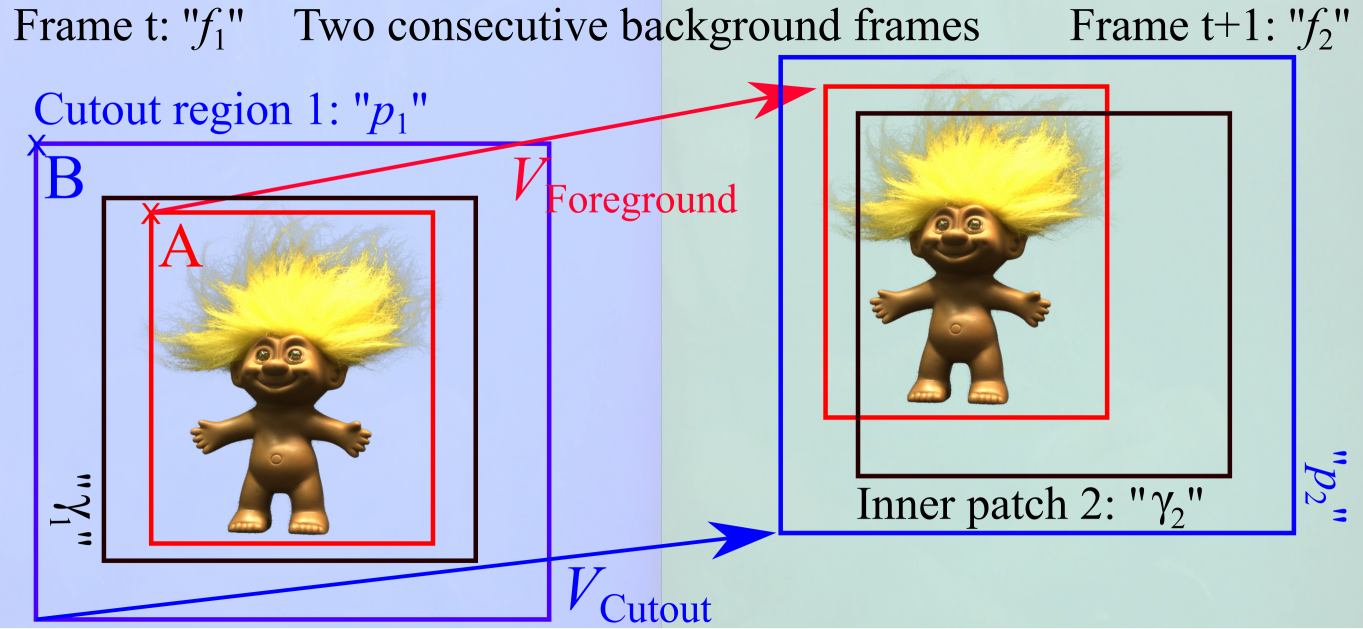}
  \end{center}
  \caption{We operate on different image sizes. Consecutive camera frames are denoted as $f_1$ and $f_2$. Our neural network operates on patches $p_1$ and $p_2$ and on inner patch regions $\gamma_1$ and $\gamma_2$. In order to train our network, we augment two input patches with simulated foreground and background movements. Given two background frames, we randomly sample a foreground position A, a cutout position B and composite the first input patch $p_1$. We displace foreground and cutout region by two randomly sampled displacement vectors $V_{Foreground}$ and $V_{Cutout}$ for the second input patch $p_2$. Similarly, we obtain our ground truth alpha matte.}
  \label{augmentation}
\end{figure}

\textbf{Training:} At the training of our network, both decoder outputs are compared to the corresponding alpha and RGB ground truth. The loss is defined as 
\begin{equation}
\begin{split}
 \mathcal{L}^{i,j}_{\alpha} &= \sqrt{(\alpha^{i,j}_p - \alpha^{i,j}_g)^2 + \epsilon^2)},\\
 \mathcal{L}^{i,j}_{c} &= \sqrt{(c^{i,j}_p - c^{i,j}_g)^2 + \epsilon^2)}, \\
 \mathcal{L}^{i} &= \omega_\alpha \mathcal{L}^{i,1}_{\alpha} +
 \omega_\alpha \mathcal{L}^{i,2}_{\alpha} \\
 &\hspace{0.5mm}+ \omega_c \mathfrak{m}^{i,1} \mathcal{L}^{i,1}_{c} + \omega_c \mathfrak{m}^{i,2} \mathcal{L}^{i,2}_{c}\\
 \end{split}
\end{equation}
in which $\mathcal{L}^i$ is the ($\omega_\alpha$,$\omega_c$)-weighted superposition of the $\alpha$-prediction loss $\mathcal{L}^{i,j}_{\alpha}$ and the color prediction loss $\mathcal{L}^{i,j}_{c}$, with $i$ indicating pixels and $j \in {1,2}$ frames. The $\alpha$-prediction loss $\mathcal{L}^{i,j}_{\alpha}$ measures the $\alpha$-value prediction $\alpha^{i,j}_p$ in comparison to the ground truth $\alpha^{i,j}_g$. Similar, the color loss $\mathcal{L}^{i,j}_{c}$ is calculated comparing the predicted color $c^{i,j}_p$ for each channel to the ground truth $c^{i,j}_g$. All 4 channels are scaled to a range of $[0,1]$. The color loss is $\mathcal{L}^{i,j}_{c}$ masked and only active for pixels where 
\begin{equation}
    \mathfrak{m}^{i,j} = 
    \begin{cases}
        0,\hspace{1mm}\textrm{if}\hspace{1mm} \alpha^{i,j}_g = 0,\\
        1,\hspace{1mm}\textrm{if}\hspace{1mm} \alpha^{i,j}_g > 0
    \end{cases}
\end{equation}
since we do not want the network to estimate foreground colors if the foreground cannot be seen. Similarly, in case of foreground movements, prediction of parts that are not visible in both input patches is prevented by limiting the output region. Thus, the overall loss $\mathcal{L}_{overall} = \sum_{i \in \gamma} \mathcal{L}^i$ is evaluated at the inner region $\gamma$ of each patch only, leaving a loss-free region of 50 pixels surrounding $\gamma$ (see Figure \ref{predictions} and \ref{augmentation}). In doing so, we ensure that each inner region pixel exists in both input frames, even if the foreground moves by up to 50 pixels.

As part of our training, we use foreground and background displacement augmentation (Figure \ref{augmentation}), simulating foreground movements and a freely moving camera. In detail, given two consecutive frames of our background dataset, we first sample a position $A$ within the background, composite a foreground, and cut out the first training input by sampling a position $B$. Then, we randomly sample two vectors $V_{Foreground}$ and $V_{Cutout}$ that displace foreground and cutout positions for the second frame $f_2$ and cut out a patch $p_2$. As before, we limit all movements to a maximum of 50 pixels. 

We follow a similar approach as Xu \etal \cite{Xu:2017} and crop image pairs to different sizes (320$\times$320, 480$\times$480, and 640$\times$640) and downscale them to patches of size 320$\times$320, thus covering multiple scales. Further training details can be found in Section \ref{details}.

\textbf{Network deployment:} As part of our image processing pipeline, we calculate foreground and alpha estimations for the consecutive frames $f_{1}$ and $f_{2}$. In the remainder of this section, we describe the network deployment or inference for the first frame $f_1$ only, which is similarly applied for the second frame $f_{2}$. We define $f_1=f$, $p_1=p$ and $\gamma_1=\gamma$. Similar to Yu \etal \cite{Yu:2020}, the incoming frame $f$ is subdivided into a set of overlapping patches $p^a, p^b, p^c, ... \in P$, which are sequentially processed by the network and we gain a set of overlapping predictions, the corresponding patches $\gamma^a, \gamma^b, \gamma^c, ... \in \Gamma$. All patches in the set $P$ are of size 320$\times$320 and all inner patches in $\Gamma$ are of size 220$\times$220. The overlap between neighboring patches in $P$ is 100 pixels so that also the inner patches $\Gamma$ overlap by 50 pixels. Two neighboring patches $\gamma^a, \gamma^b \in \Gamma$ are linearly blended in the overlapping area $\gamma^a \cup \gamma^b$, with blending weights proportional to the distances to the patch boundaries (see \cite{Yu:2020}). Thus the influence of the prediction $\gamma^a$ gradually diminishes at its boundaries with the increasing influence of the neighboring prediction $\gamma^b$. 

Following this overlap-blend approach, we obtain full-size foreground predictions with colors $C_{fg, pred}$, and alphas $\alpha_{pred}$ for each frame $f$. For the composition of the foreground prediction with a new background $C_{bg}$, i.e.\ VFX content, we modify the original matting equation \ref{mattingequation} to
\begin{equation} 
    \begin{split}
  C &= \alpha_{pred} (\alpha_{pred} C_{fg, pred} + (1-\alpha_{pred}) C_{fg, orig}) \\
  &+ (1-\alpha_{pred}) C_{bg}.
  \end{split}
  \label{newmattingequation}
\end{equation}
Thus, we blend the color information, using the original colors $C_{fg, orig} = f$ if $\alpha_{pred}$ is close to 1 and the predicted color $C_{fg, pred}$ if $\alpha_{pred}$ is close to 0. In short, equation \ref{newmattingequation} introduces our color spill correction.  

\subsection{Implementation details}
\label{details}
In the following, we want to add further details on training and implementation of our method and our implementation of DIM \cite{Xu:2017} to facilitate reimplementations. These details are described detached from the description of the method in order to increase readability of the previous subsection.

\textbf{Architecture:}
The architecture of our network can be seen in Figure \ref{network}. Input dimensions are 6$\times$320$\times$320 and output dimensions are 4$\times$320$\times$320, for each decoder. The encoder is similar to Xu \etal \cite{Xu:2017} and consists of blocks with 2 or 3 2D convolutional layers (kernel size 3), followed by group normalization (as introduced by Wu and He \cite{Wu:2020}), rectified linear unit (ReLU) activation and max pooling. With each encoder block, the output shape is reduced by a factor of 2, while the number of channels increases in the sequence 64, 128, 256, 512 to 1024. The encoder parameters of the 14 layers are initialized by the pre-trained VGG16 network of Simonyan and Zisserman \cite{Simonyan:2015}. Both decoders have the same architecture with blocks consisting of convolutional layers followed by group normalizations, ReLUs, and 2D transposed convolutions (kernel size 6 and stride 2). The final convolution (yellow in Figure \ref{network}) is followed by a ReLU activation and clipping of values to a maximum of 1. This is motivated by the observation that color values tend to be evenly distributed. This is in contrast to alphas values, which are typically close to one or close to zero, for which sigmoid activation is preferred. The decoder weights are initialized randomly. Similar to Forte and Piti\'{e} \cite{Forte:2020}, we use a mini-batch size of 1. The long skip connections (Figure \ref{network}), connecting encoder and decoder, are motivated by U-nets introduced by Ronneberger \etal \cite{Ronneberger:2015} and which are beneficial for the prediction of fine-grained details. In our implementation, the skip connections are additive as in ResNets introduced by He \etal\cite{He:2015}. Our dual decoder network consists of 81.59M trainable parameters.

\textbf{Parameters:}
In Equation 2, $\omega_c$ is set to 0.5 and $\omega_\alpha$ to 1. As described in the paper, the color loss $\mathcal{L}^{i,j}_{c}$ can be masked by $\mathfrak{m}^{i,j}$. If it is not masked, meaning $\mathfrak{m}^{i,j} \neq 0$, color prediction errors have a larger impact on the loss $\mathcal{L}^i$ than alpha prediction errors, since three channels are each weighted by $\omega_c = 0.5$. In Equation 2, $\epsilon$ is set to $1e-6$. In Figure \ref{augmentation}, each displacement vector $V_{Foreground}$ and $V_{Background}$ is limited to 50 pixels, resulting in a maximal combined foreground-background movement of 100 pixels.

\textbf{Training:}
As training dataset we use the 432 foreground images of the Adobe dataset \cite{Xu:2017} and 5000 frames of recorded background samples from our demonstrator setup. Foreground and background samples are divided into training (80\%) and validation (20\%) data each. Within each dataset (training and validation) foreground and background combinations are sampled resulting in a training set of 34560 and a validation set of 8640 foreground-background tuples. The data split and the combinations were created once and stayed the same during training. A tuple consists of one foreground image and two consecutive background images with different background colors. During training, within one foreground-background tuple, random cropping augmentation is conducted as illustrated in Figure \ref{augmentation}. This randomized real-time augmentation leads to increased data diversity. In addition, augmentation techniques such as randomly flipping, changing contrasts, or adding color jitter are applied. Our network is implemented in pytorch. As optimizer we chose stochastic gradient descent with an initial learning rate of 0.01 and momentum of 0.9. We enforce a steadily falling validation loss by reinitialization with a saved checkpoint from the previous epoch, if the validation loss increase. If no decrease in validation loss is achieved within 2 epochs, the learning rate is decreased by a factor of 0.6. If no decrease in validation loss is achieved within 5 epochs, training is terminated. The network was trained for 50 epochs, which took 89 hours using an Nvidia GeForce RTX 2080 Ti GPU.

\textbf{Inference:}
During training, we apply a randomized movement augmentation to simulate foreground and background movements, to increase the robustness of the network for moving scenes. During inference, meaning the deployment of the network as part of the proposed matting system, overlapping patches are cropped from the camera frames $f_1$ and $f_2$ in a fixed grid and corresponding patches $p_1$ and $p_2$ are extracted from the same coordinates but from two consecutive frames, meaning that using a camera speed of @\unit[100]{fps} the second frame $f_2$ was captured \unit[10]{ms} after $f_1$. Thus, moving or non-static foregrounds and backgrounds have a displacement in $p_1$ and $p_2$. 

\section{Results}
\label{sec:results}
\textbf{Algorithms:} We compare our method to the following related algorithms. The first method is denoted \textit{BSM}, which is our implementation of Smith and Blinn's \cite{Smith:1996} blue screen matting. The authors of BSM claim that matting with a ``multi-background technique'' can only be applied in a static case, without ``live actors or other moving objects''. In addition BSM needs a perfect knowledge of the background colors. Furthermore, Smith and Blinn write that an application without these requirements is ``powerful''. We see our approach as an extension of their work, which is able to overcome these limitations and show that multi-background matting can be applied on dynamic scenes. 
\begin{figure}[t]
\begin{center}
 \includegraphics[width=0.99\linewidth]{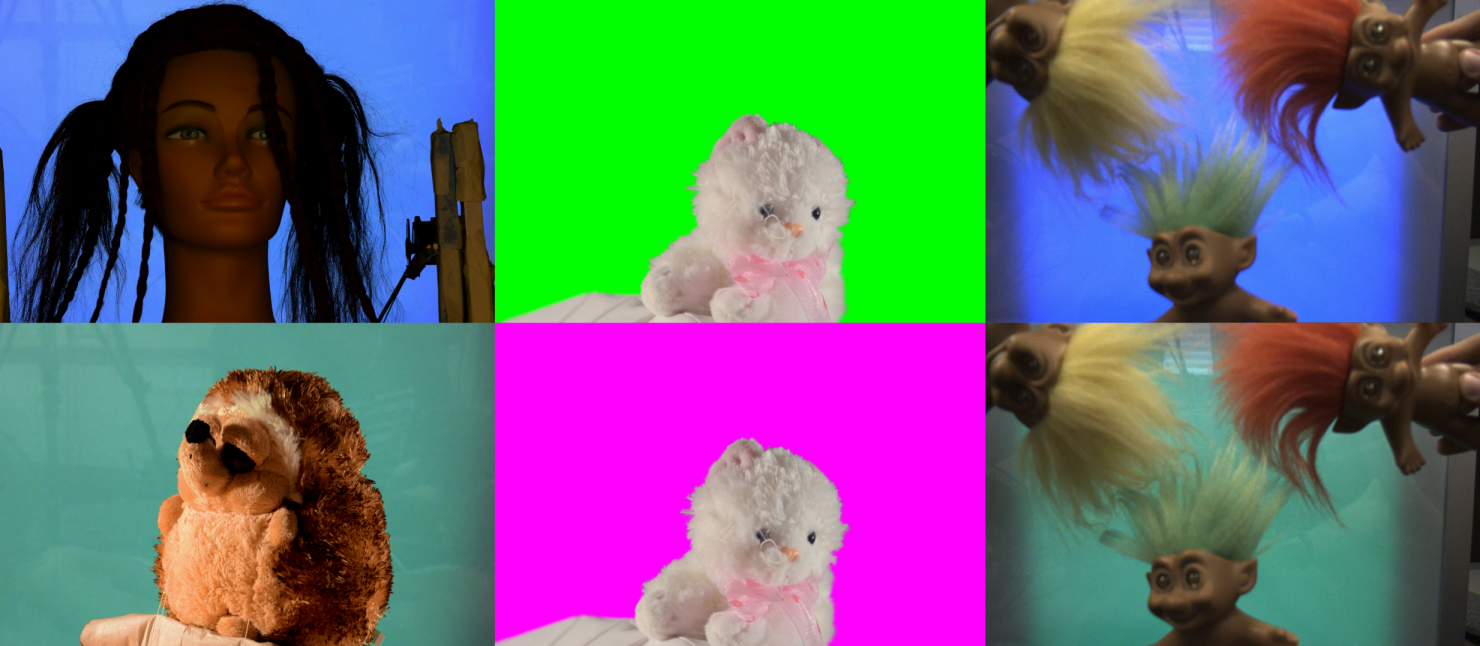}
 \end{center}
 \caption{Evaluation dataset: (left) Composition of \textit{Castle} and \textit{Dmitriy} of Erofeev \etal's \cite{Erofeev:2015} benchmark\textsuperscript{\ref{note2}} in front of a realistic blue and green background respectively, (middle) composition of \textit{Alex} (also \cite{Erofeev:2015}) in front of a pure and orthogonal green and purple background and (right) two consecutive frames with \textit{Trolls} using our demonstrator.}
 \label{test_dataset}
\end{figure}

The second method \textit{CIM} is our implementation of Grundh\"{o}fer \etal's \cite{Grundhoefer:2010} method, in which they use hardware keying units (\textit{Ultimatte 11}) with additional Bayesian matting. Since we do not own Ultimatte 11, we cannot recreate this particular matting pipeline. Instead, CIM uses ground truth alpha mattes or ``learning-based digital matting'' by Zheng and Kambhamettu \cite{Zheng:2009} on the \textit{Troll} sequence, for which no ground truth is available. CIM requires complementary backing colors, that sum up to a neutral white, which is the core idea of their color spill neutralization. Therefore, similar to BSM, perfect backgrounds are needed, however, this requirement is only fulfilled by the pure color background sequences.

In contrast to the previous two alternating backdrop methods, the following two methods are trimap-based. In our scenario trimaps are generated from the ground truth alpha matte by setting all pixels that fulfill $ 0 < \alpha < 1$ to ``unknown'', followed by morphological dilation of this gray zone. The distribution of foreground, background and unknown areas of the test sequences can be seen in Table \ref{table2}.

The third matting technique \textit{DIM} is an implementation of Xu \etal \cite{Xu:2017}. The basic structure of this neural network can be explained with the help of Figure \ref{network}, applying a few modifications. Instead of the second input patch $p_2$, the network receives a trimap and the second decoder branch is omitted. By comparing to DIM we can directly measure any performance gain achieved by substituting the trimap version with our dual-frame version.

The forth method Semantic Image Matting or \textit{SIM} by Sun \etal \cite{Sun:2021}, we use official repository, enhances the matting results by first classifying the foreground, creating a semantic trimap which guides the matting network. In our evaluation the classification into the classes \textit{fur}, \textit{hair\_hard} and \textit{hair\_easy} should be beneficial along with \textit{motion} for the Trolls sequence. The class \textit{defocus} may help to cope with chromatic aberration. SIM ranks in the top-5 of Rhemann\footnote{\url{alphamatting.com}} \etal \cite{Rhemann:2009} benchmark.

Finally, we compare with \textit{FBA}, the official demonstrator of Forte \etal \cite{Forte:2020}, which is publicly available. The method is currently the leading algorithm on the benchmark\footnote{\label{note2}\url{videomatting.com}} of Erofeev \etal \cite{Erofeev:2015} and represents the state-of-the-art in trimap-based matting.
\begin{table}
\begin{tabular}[h]{c||c|c|c|c}
     & Alex & Dmitriy & Castle & Troll  \\
    \hline\hline
    White area in trimap & 18.35\% & 23.81\% & 29.66\% & 29.87\% \\
    Black area in trimap & 71.35\% & 65.75\% & 23.50\% & 31.13\% \\
    Gray area in trimap & 10.29\% & 10.44\% & 46.84\% & 39.01\% \\
    \hline
    max(PSNR) & 60.0 & 60.0 & 60.0 & n.a. \\
    max(VMAF) & 99.86 & 99.98 & 97.43 & n.a.\\
    max(MS-SSIM) & 0.9999 & 0.9999 & 0.9999 & n.a.\\
\end{tabular}\vspace{1mm}
\caption{Dataset properties. Top rows: trimap color distribution with the percentage of foreground (white pixel), background (black pixel) and unknown region (gray pixel). Bottom rows: upper metric limits for PSNR, MS-SSIM and VMAF.}
\label{table2}
\end{table}

\textbf{Test dataset and metrics:} We evaluate our method on the three sequences from Erofeev \etal \cite{Erofeev:2015}, called \textit{Dmitriy}, \textit{Alex} and \textit{Castle}. While this dataset includes more sequences, these are the only sequences for which ground truth alpha mattes are publicly available. In this paper, these are necessary for the alternating color background sequence generation and for evaluation. The foregrounds Dmitriy, Alex and Castle are composited with alternating blue and green backdrops acquired by our demonstrator or with complementary pure green and pure purple backings. The resulting samples can be seen in Figure \ref{test_dataset} and the top row of Figure \ref{fig:comparison}. Evaluation on composition level has several advantages compared to independent evaluation of alpha values and foreground color prediction, since the effect of errors in foreground color prediction are linked to alpha values. If alpha is zero, colors may be erroneous without impacting the composition. Evaluation on the composition solves this issue by measuring alpha and color estimation at the same time. Furthermore, some measures, such as Gradient and Connectivity (see below), do not make sense for colors. 

For quantitative evaluation, the matting results are composited with checkerboard backgrounds, compared to the ground truth, and evaluated with the following metrics: PSNR, MS-SSIM \cite{Wang:2003} and the perceptual Video Multi-Method Assessment Fusion (VMAF) \cite{li2016toward}. The peak-signal-to-noise ratio (PSNR) is based upon the mean squared error (MSE), which measures a pixel-wise comparison to the ground truth. Temporal inconsistencies or flickering are of major importance, since the human visual system is strongly affected by them. Furthermore, error concealment methods seldomly correct matting results, but diminish the impact of occurring errors. This is why we also evaluate the structural similarity (MS-SSIM) and perceptual (VMAF) scores. We calculate all metrics using FFmpeg \cite{tomar:2006} and lossless H.264 encoding ($qp=0$) with $YCbCr = \text{4:4:4}$, meaning without chroma subsampling. The highest achievable scores can be obtained by comparing the ground truth to itself. These upper metric limits can be found in Table \ref{table2} and it can be seen that the maximal VMAF score is dataset dependent. The metrics PSNR, MS-SSIM and VMAF are frequently used in the video coding community that has a long record of in-depth video quality assessment.

In the matting community, the alpha mattes are typically evaluated independently from the foreground colors. In Table \ref{table1}, we show results on the alpha prediction measures SAD, MSE, Gradient and Connectivity as proposed by Rhemann \etal \cite{Rhemann:2009}. The results of our method ours$_{ma}$ on these scores are on a similar level as FBA and SIM and far better than DIM, BSM and CIM. Nevertheless, this paper focuses on foreground color prediction as a part of a matting system and thus the results on the scores MS-SSIM, PSNR and VMAF are of major importance.

As part of our assessment, we provide quantitative results on our \textit{Troll} sequence, a video captured by our demonstrator. In contrast to the virtual dataset, the Troll dataset has additional challenges such as noise, chromatic aberration, and non-homogeneous, not fully known backgrounds. The \textit{Troll} dataset can only be used for qualitative evaluation as no ground truth exists. The Troll dataset and other sequences from the demonstrator are publicly available at \url{(anonymous_submission)}. 

\textbf{Discussion:} Quantitative results of seven algorithms on our test dataset with real-world backgrounds can be found in Table \ref{table1} and the following five key findings can be observed.

First, our method ours$_{ma}$ performs better than BSM \cite{Smith:1996} and CIM \cite{Grundhoefer:2010} by a large margin. This comparison is important as these three methods receive the same input data and can share the same hardware setup.

Second, the comparison of DIM \cite{Xu:2017} to our method shows the direct benefit of replacing the trimap with a 2 frame input, since both networks are otherwise similar in architecture and trained on the same dataset with identical hyper parameters. The results show a drastic gain in performance, which is directly linked to the architecture changes introduced with our method.

Third, our method performs on a similar level as FBA by Forte \etal \cite{Forte:2020} and SIM by Sun \etal \cite{Sun:2021}, which proofs that ours$_{ma}$ can provide state-of-the-art results. Note that FBA and SIM need a trimap as input which is not easily obtained and any errors in the foreground and background areas of the trimap directly lead to errors in the alpha mattes. In our evalutation, the trimaps given to FBA and SIM are without errors and contain 53.16\% to 89.71\% of ground truth data (white and black area in Table \ref{table2}). 
\begin{figure}[t]
  \begin{center}
    \includegraphics[width=\linewidth]{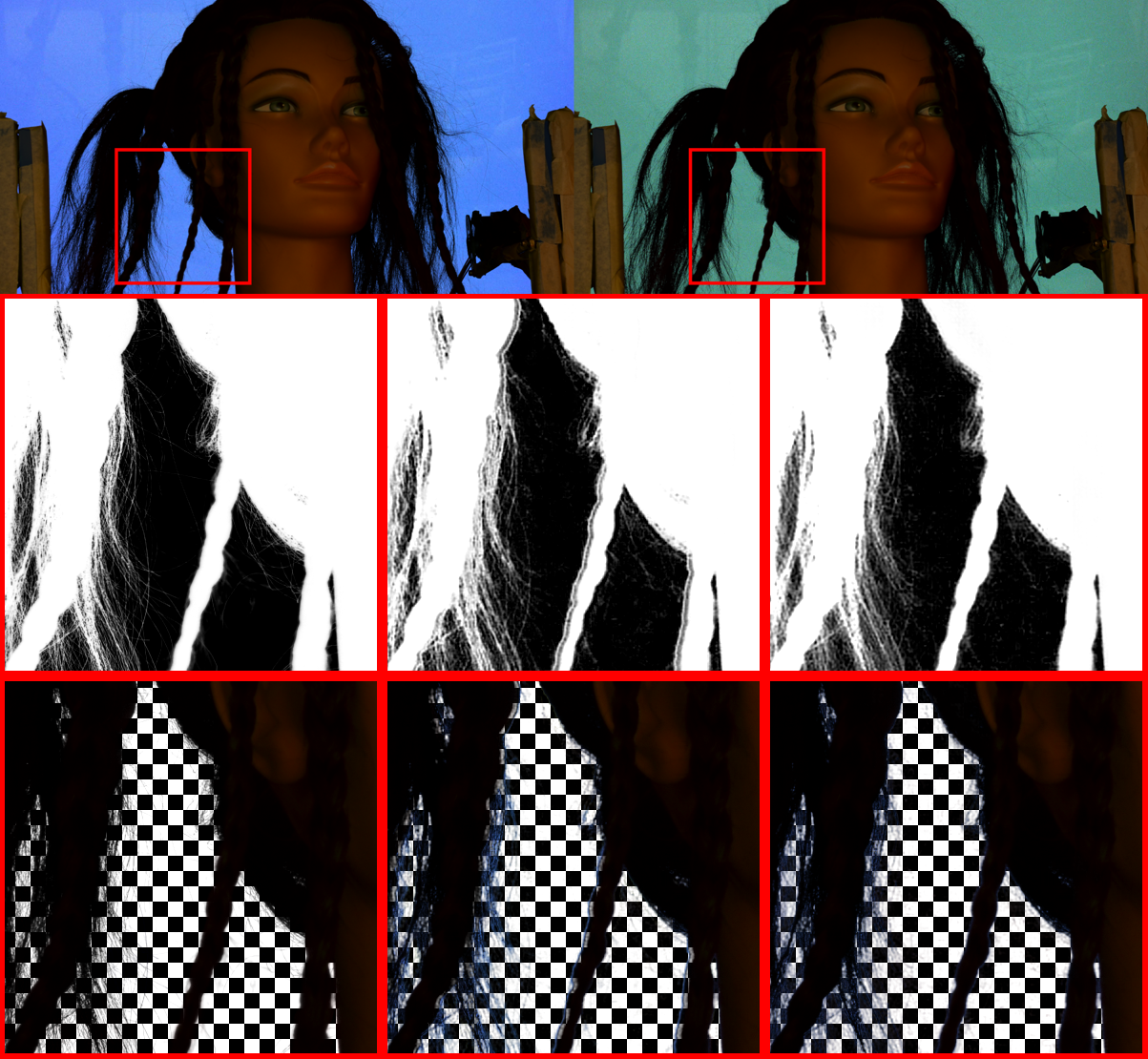}
  \end{center}
  \caption{Effect of motion augmentation illustrated on the Castle sequence: (top): input frames to the network and overview of the cropped region, (lower, left): ground truth alpha matte and ground truth superposition with a checkerboard background corresponding to the blue background input image (lower, center): predictions by the proposed network ours$_{sa}$ trained without motion augmentation or static and (lower, right): predictions by the proposed network ours$_{ma}$ trained using motion augmentation. The lack of motion augmentation during training leads to double contours and increases color spill.}
  \label{motionaugmentation}
\end{figure}

Fourth, an ablation study illustrates the impact of our motion augmentation. The column ours$_{sa}$ shows results of our network trained statically, while ours$_{ma}$ used motion augmentation during training. Static means that $V_{Foreground} = 0$ and $V_{Cutout} = 0$ (cf.~Figure \ref{augmentation}). From all scores, it can be observed that the motion augmentation is of enormous importance. In Figure \ref{motionaugmentation} both models show qualitative results on two frames of the non-static Castle dataset. The static model ours$_{sa}$ clearly fails to cope with foreground movements, which leads to double contours in the alpha matte and increased color spill in the composition with a checkerboard background.
\begin{table*}
\resizebox{0.88\linewidth}{!}{
\centering
\begin{tabular}[h]{c|c|c||c|c|c||c|c|c|c}
    Data & Metric & better & DIM$^6$ \cite{Xu:2017} & FBA \cite{Forte:2020} & SIM \cite{Sun:2021} & BSM$^6$ \cite{Smith:1996} & CIM$^6$ \cite{Grundhoefer:2010} & ours$_{sa}$ & ours$_{ma}$ \\
    \hline\hline
    Trimap & & & \cmark & \cmark & \cmark & - & - & - & -\\ 
    2 frames & &  & - & - & -  & \cmark & \cmark & \cmark & \cmark \\
    \hline
    \hline
    Alex    & MS-SSIM & $\uparrow$ & 0.98332 & \textbf{0.99691} & 0.99544 & 0.93664 & 0.96418 & 0.95779 & \textbf{\underline{0.99714}}\\
    Dmitriy & MS-SSIM & $\uparrow$ & 0.98984 & 0.99566 & \textbf{0.99574} & 0.89001 & 0.92781 & 0.96851 & \textbf{\underline{0.99574}}\\
    Castle & MS-SSIM & $\uparrow$ & 0.98327 & \textbf{\underline{0.98847}} & 0.98675 & 0.96927 & 0.97122 & 0.97759 & \textbf{0.98705} \\
    \hline
    Alex    & PSNR & $\uparrow$ & 31.311 & \textbf{43.374} & 41.681 & 26.367 & 29.655 & 30.931 & \textbf{\underline{44.867}} \\
    Dmitriy & PSNR & $\uparrow$ & 37.805 & \textbf{\underline{39.852}} & 39.678 & 24.194 & 27.319 & 29.623 & \textbf{39.482} \\
    Castle & PSNR & $\uparrow$ &27.811 & \textbf{\underline{33.088}} & 32.814 & 27.544 & 27.805 & 29.346 & \textbf{31.759} \\
    \hline
    Alex    & VMAF & $\uparrow$ & 94.689 & \textbf{99.370} & 99.164 & 74.482 & 92.388 & 93.585 & \textbf{\underline{99.513}} \\
    Dmitriy & VMAF & $\uparrow$ & 99.553 & \textbf{\underline{99.807}} & 99.572 & 67.088 & 83.200 & 92.782 & \textbf{99.785} \\
    Castle & VMAF & $\uparrow$ & 82.543 & \textbf{\underline{88.677}} & 85.411 & 69.725 & 72.663 & 79.461 & \textbf{87.314} \\    
    \hline
    \hline
    Alex    & SAD & $\downarrow$ & 5.251 & \textbf{2.382} & 2.571 & 10.841 & 6.125 & 9.620 & \textbf{\underline{1.198}} \\
    Dmitriy & SAD & $\downarrow$ & 6.966 & 1.496 & \textbf{1.461} & 14.789 & 9.298 & 10.992 & \textbf{\underline{1.331}} \\
    Castle & SAD & $\downarrow$ & 28.057 & \textbf{\underline{3.916}} & 6.302 & 50.792 & 18.620 & 22.795 & \textbf{10.780} \\
    \hline
    Alex    & MSE ($10^3$) & $\downarrow$ & 3.192 & \textbf{1.151} & 2.371 & 22.090 & 25.297 & 27.231 & \textbf{\underline{0.452}} \\
    Dmitriy & MSE ($10^3$) & $\downarrow$ & 9.129 & 0.898 & \textbf{\underline{0.745}} & 48.158 & 43.058 & 33.835 & \textbf{1.084} \\
    Castle & MSE ($10^3$) & $\downarrow$ & 4.310 & \textbf{\underline{0.188}} & 0.406 & 31.412 & 11.084 & 7.410 & \textbf{1.273} \\
    \hline
    Alex    & Grad ($10^{-1}$) & $\downarrow$ & 93.36 & \textbf{30.79} & 39.29 & 683.53 & 877.07 & 765.106 & \textbf{\underline{18.29}} \\
    Dmitriy & Grad ($10^{-1}$) & $\downarrow$ & 485.13 & \textbf{\underline{45.68}} & 54.49 & 2422.75 & 2248.97 & 1740.41 & \textbf{91.46} \\
    Castle & Grad ($10^{-1}$) & $\downarrow$ & 381.44 & \textbf{\underline{53.03}} & 79.32 & 7896.57 & 3594.16 & 2483.34 & \textbf{242.89} \\
    \hline
    Alex    & Conn ($10^{-2}$) & $\downarrow$ & 20.148 & \textbf{9.369} & 12.730 & 80.989 & 61.996 & 97.17 & \textbf{\underline{4.132}} \\
    Dmitriy & Conn ($10^{-2}$) & $\downarrow$ & 51.299 & 8.431 & \textbf{\underline{7.404}} & 128.131 & 94.101 & 110.85 & \textbf{10.760} \\
    Castle & Conn ($10^{-2}$) & $\downarrow$ & 157.124 & \textbf{\underline{12.545}} & 26.446 & 459.208 & 184.359 & 211.11 & \textbf{71.161} \\    
    \hline
    \hline
    Trolls& MAD & $\downarrow$ & 2.5319 & \textbf{1.0696} & 1.2605 & 1.0284 & 1.1385 & 0.9393 & \textbf{\underline{0.8866}} \\
\end{tabular}
}
\vspace{1mm}
\caption{Seven matting algorithms are evaluated on three ground truth datasets: Alex, Dmitriy, and Castle which are composited with real-world backgrounds as can be seen in the left columns of Figure \ref{fig:comparison} and on the Trolls sequence captured by our demonstrator (right columns in Figure \ref{fig:comparison}). The best results of the two groups, using a trimap or using 2 frames as input, are marked in bold and the best results globally are underlined. There are three experiments: (1) evaluation is done on composition level with a checkerboard background. Here alpha and predictions are jointly measured using MS-SSIM, PSNR and VMAF scores. (2) alpha predictions are measured independently using SAD, MSE, gradient and connectivity. (3) there is no ground truth for the demonstrator sequence Trolls. However, the mean alpha deviation (MAD), as in Eq. \ref{MAD}, measures alpha value consistency across the predicted video sequence. The experiments show that FBA \cite{Forte:2020}, SIM \cite{Sun:2021} and our algorithm perform similarly and achieve top scores. The two columns on the right show a short ablation study on the impact of our motion augmentation ours$_{ma}$ during training versus a static version ours$_{sa}$. 
}
\label{table1}
\end{table*}

Fifth, Table \ref{table1} provides MAD scores on the Troll sequence, for which no ground truth alpha values are available. However, we can measure temporal alpha value consistency as follows. With changing background colors the average alpha values of the foreground should remain unchanged. The accumulated mean alpha value deviation
\begin{equation}
    MAD = \frac{1}{N} \sum^{N-1}_{n=1} \frac{1}{wh}\left |  \sum_{x=1}^{w}\sum_{y=1}^{h}\alpha_n(x,y) - \sum_{x=1}^{w}\sum_{y=1}^{h}\alpha_{n+1}(x,y) \right | 
    \label{MAD}
\end{equation}
measures alpha estimation inconsistencies between frames on the Troll dataset, with $N = 25$ the number of frames, alpha values ranging from 0 to 255 and a video resolution with width $w=2448$ and height $h=1600$. In theory, MAD may also consist of an alpha deviation due to foreground movements and deformations. However, this share of the MAD score is the same for all algorithms. Table \ref{table1} illustrates that the alpha predictions of ours$_{ma}$ achieve the best MAD score and are thus most consistent.

As a sanity check we evaluate on an artificial dataset with pure green and purple backgrounds (see center columns in Figure \ref{test_dataset}). As we expect, BSM performs best and creates perfect results on Dmitriy and Alex (see Table \ref{table2}). Only the Castle sequence scores are non-perfect with MS-SSIM 0.9993, PSNR 49.90 and VMAF 97.20, which is probably due to numerical inaccuracies.

Qualitative results on our test dataset with real-world backgrounds and the Troll sequence can be found in Figure \ref{fig:comparison}. In the first three columns, it can be seen that the foreground movement has a drastic impact on BSM and CIM, which show color seams and erroneous transparencies in the foreground. It can be seen on the lower row of CIM's Castle examples that superimposing alpha mattes without registration leads to visible double contours at single hair strands. On the right column, in the Troll sequence, we observe that DIM, FBA, SIM and CIM suffer from blue color spill, while our method shows close to no color spill, a slight blue shade in the green hair merely. Quite curiously, BSM seems to have shades of red in the yellow hairs and yellow shades in the green hairs, which seems to be the triangulation error. Furthermore, FBA, SIM and CIM seem to be affected by the noise in this dataset. 

A comparison of the consecutive frames from the Troll dataset, as in Figure \ref{aberration2}, \ref{test_dataset} and \ref{fig:comparison}, reveals that single hair fibers can be more easily recognized with green backing, which is most probably due to chromatic aberration \cite{Korneliussen:2014}. As can be seen in Figure \ref{aberration2}, this effect leads to overestimated alpha values for green and underestimated alpha values for blue backdrops, which we experience systematically with all seven matting algorithms. While the effect of chromatic aberration as described in Figure \ref{aberration1} and \ref{aberration2} is not explicitly modeled in our approach, the joint, temporally mixed latent space in our one-encoder-dual-decoder in combination with the motion augmentation seems to increase temporal consistency. Although this is speculative, we believe that the effect of aberration and foreground motion are related in our feature space. Since neural networks interpolate and create local smoothness between known data, this could explain that robustness with respect to motion also has a positive effect on aberration compensation.\color{white}\footnote{reimplementation}\color{black}
\section{Conclusion}
\label{sec:conclusion} This paper presents a novel neural network-based dual-backdrop duplex matting system that creates high-quality alpha as well as foreground color predictions. It is temporally consistent, unaffected by noise, and shows superior color spill compensation. We compare our approach to a trimap-guided twin method that is trained and tested on the same datasets. In this experiment we clearly show that temporal backdrop duplex matting achieves superior results to the trimap-based approach. In addition, we propose a hardware set, that is actor friendly and can potentially be used in upcoming LED video wall production studios.

In the future, we intend to research the impact of chromatic aberration in more detail. This effect could be explicitly modeled in the dataset generation and augmentation, in order to further increase temporal consistency in alpha matting. Furthermore, we will investigate adaptation to view-dependent variations of the backings by camera motion prediction as in Dockhorn and Kruse \cite{DocKru2020b}.

\section*{ACKNOWLEDGMENTS}
\label{sec:acknowledgements}
This work was supported by the Federal Ministry of Education and Research (BMBF), Germany under the project LeibnizKILabor (grant no. 01DD20003) and the AI service center KISSKI (grant no. 01IS22093C), the Center for Digital Innovations (ZDIN) and the Deutsche Forschungsgemeinschaft (DFG) under Germany’s Excellence Strategy within the Cluster of Excellence PhoenixD (EXC 2122). 

\clearpage
\newcommand{\mywidth}{0.17}
\newcommand{\myheight}{0.109}
\begin{figure*}[ht]
  \begin{center}
  \rotatebox{90}{\centering{\color{white}Xg...\color{black}Sequence}}
  \begin{subfigure}
    {
    \centering
    \includegraphics[height=\myheight\textheight]{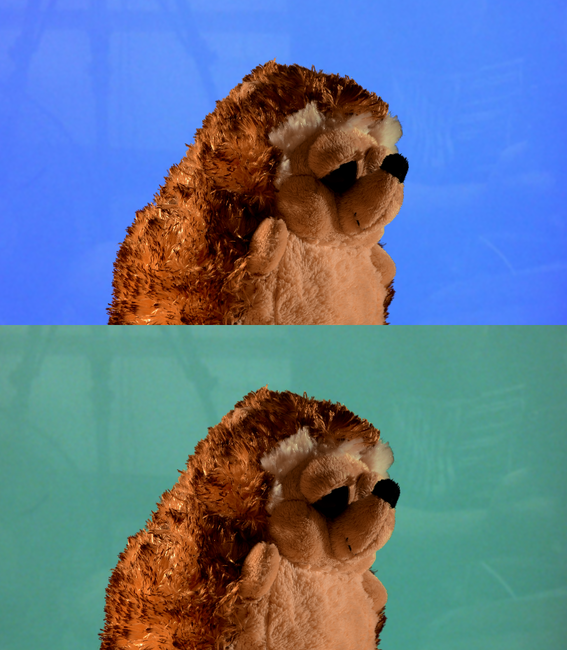}
    }	    
  \end{subfigure}
  \begin{subfigure}
    {
    \centering
    \includegraphics[height=\myheight\textheight]{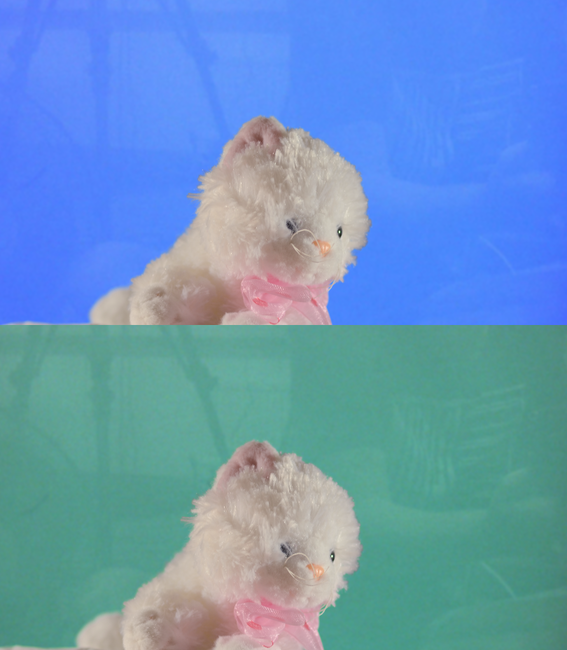}
    }
  \end{subfigure}
  \begin{subfigure}
    {
    \centering
    \includegraphics[height=\myheight\textheight]{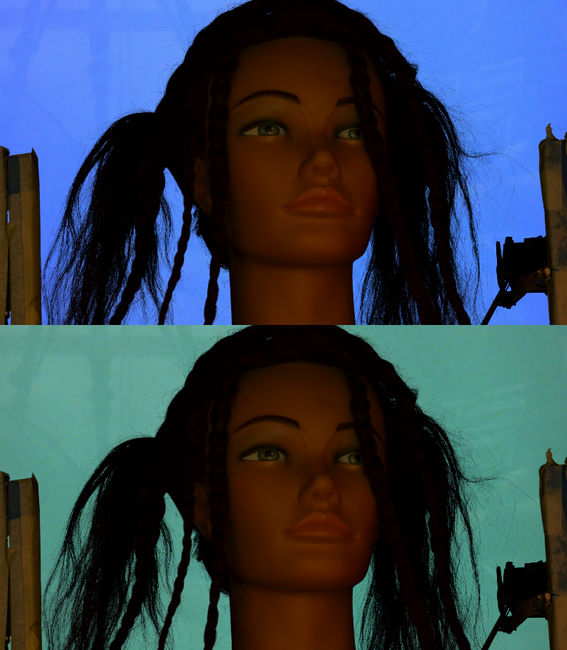}
    }
  \end{subfigure} 
  \begin{subfigure}
    {
    \centering
    \includegraphics[height=\myheight\textheight]{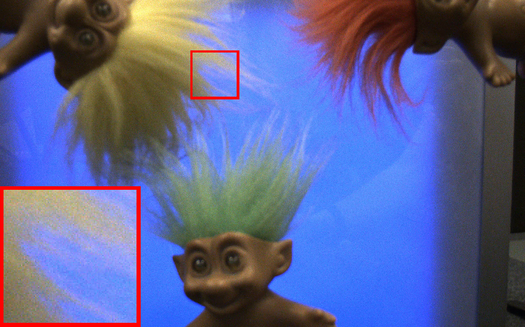}
    }
  \end{subfigure}  
  \hspace{-4mm}
  \begin{subfigure}
    {
    \centering
    \includegraphics[height=\myheight\textheight]{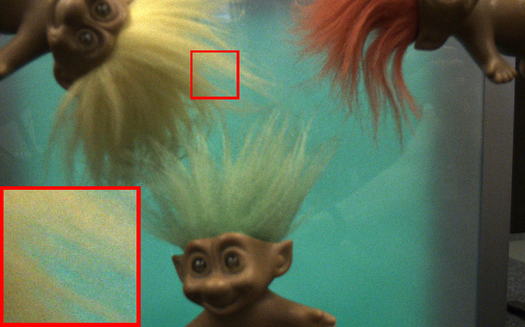}
    }
  \end{subfigure} 
  \\[-2mm]
  \rotatebox{90}{\centering{\color{white}Xg\color{black}Ground truth}}
  \begin{subfigure}
    {
    \centering
    \includegraphics[height=\myheight\textheight]{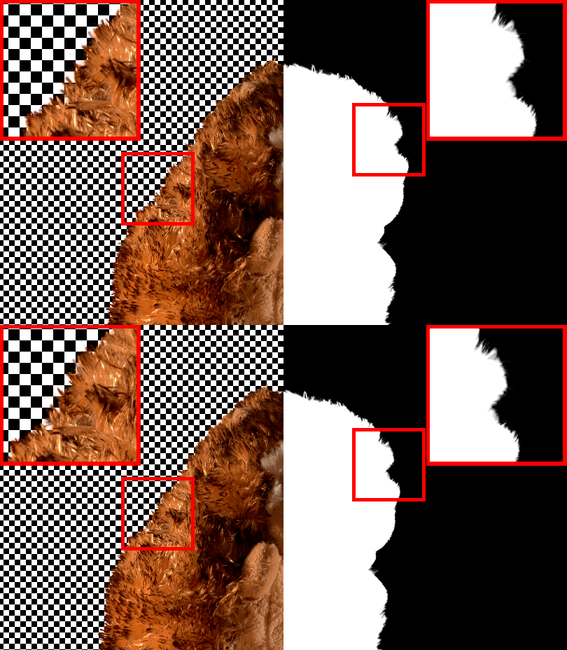}
    }	    
  \end{subfigure}
  \begin{subfigure}
    {
    \centering
    \includegraphics[height=\myheight\textheight]{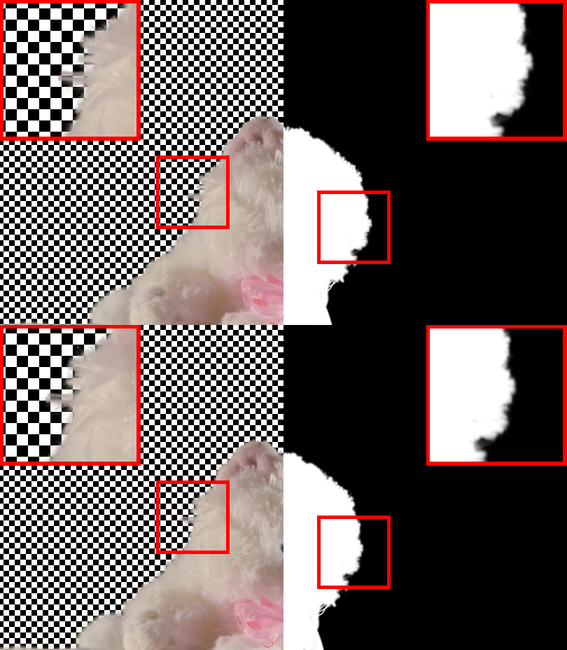}
    }
  \end{subfigure}
  \begin{subfigure}
    {
    \centering
    \includegraphics[height=\myheight\textheight]{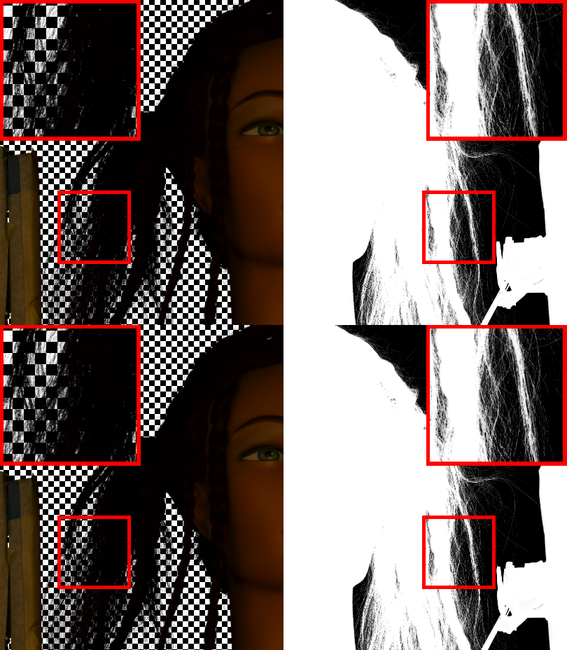}
    }
  \end{subfigure}
  \begin{subfigure}
    {
    \centering
    \includegraphics[height=\myheight\textheight]{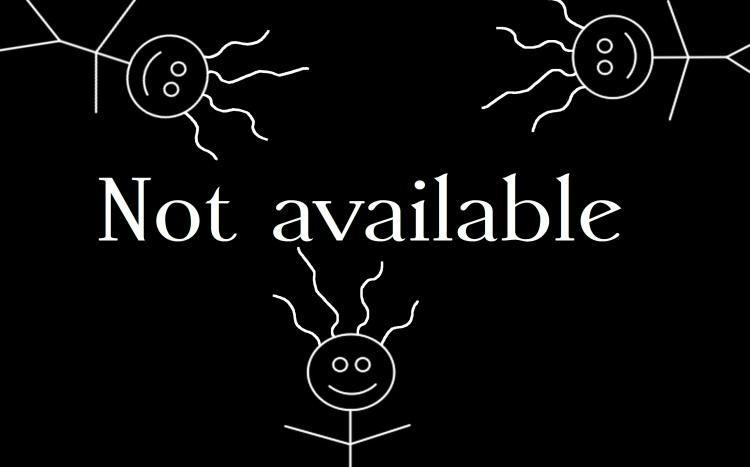}
    }
  \end{subfigure} 
  \hspace{-4mm}
  \begin{subfigure}
    {
    \centering
    \includegraphics[height=\myheight\textheight]{images/TrollGT/TrollParty_GT_small}
    }
  \end{subfigure} 
  \\[-2mm]
  \rotatebox{90}{\centering{\color{white}Xg....\color{black}DIM$^6$\cite{Xu:2017}}}
  \begin{subfigure}
    {
    \centering
    \includegraphics[height=\myheight\textheight]{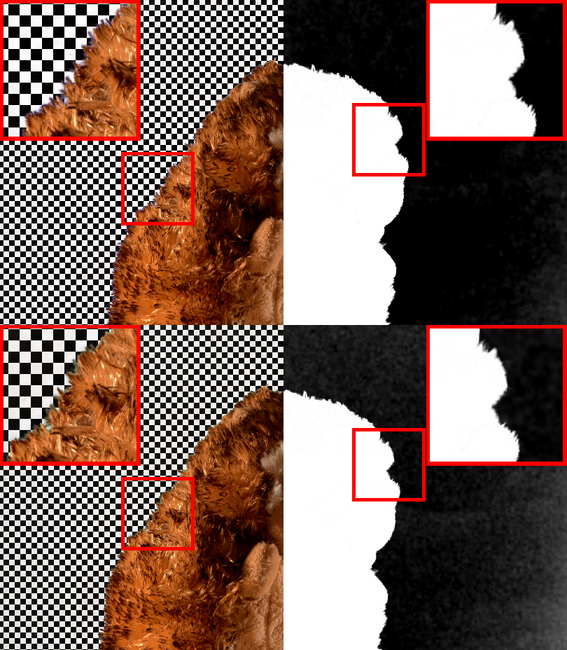}
    }	    
  \end{subfigure}
  \begin{subfigure}
    {
    \centering
    \includegraphics[height=\myheight\textheight]{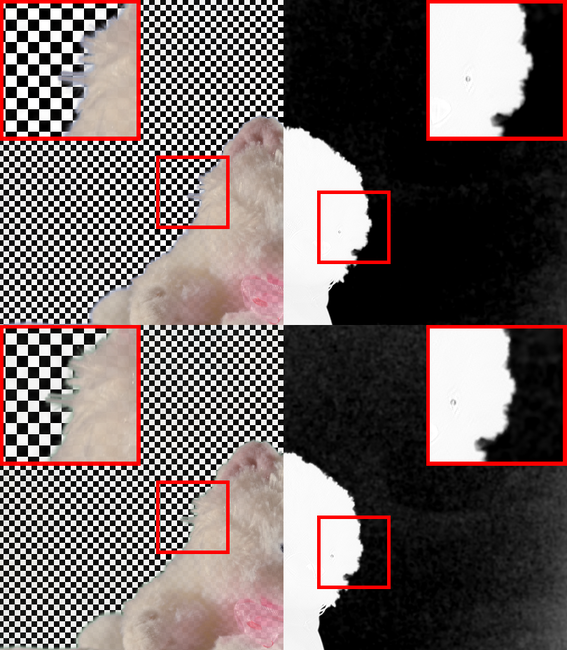}
    }
  \end{subfigure}
  \begin{subfigure}
    {
    \centering
    \includegraphics[height=\myheight\textheight]{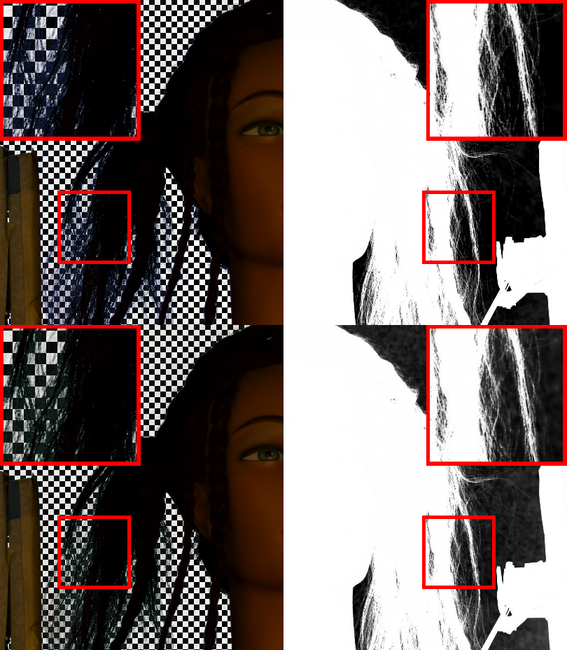}
    }
  \end{subfigure}
  \begin{subfigure}
    {
    \centering
    \includegraphics[height=\myheight\textheight]{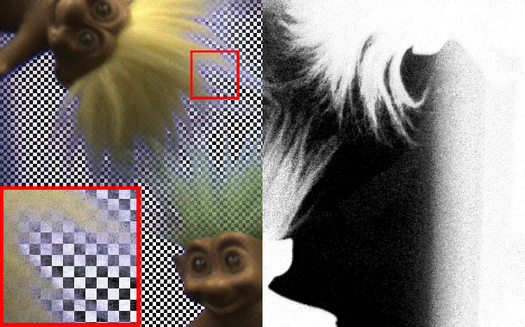}
    }
  \end{subfigure} 
  \hspace{-4mm}
  \begin{subfigure}
    {
    \centering
    \includegraphics[height=\myheight\textheight]{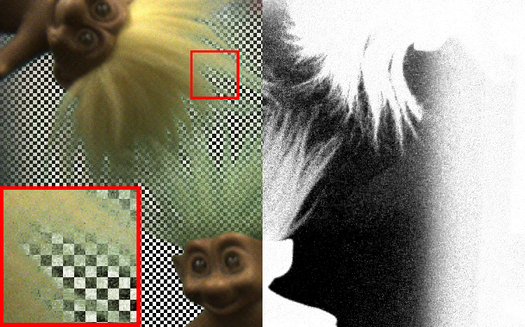}
    }
  \end{subfigure}  
    \\[-2mm]
  \rotatebox{90}{\centering{\color{white}Xg....\color{black}FBA \cite{Forte:2020}}}
  \begin{subfigure}
    {
    \centering
    \includegraphics[height=\myheight\textheight]{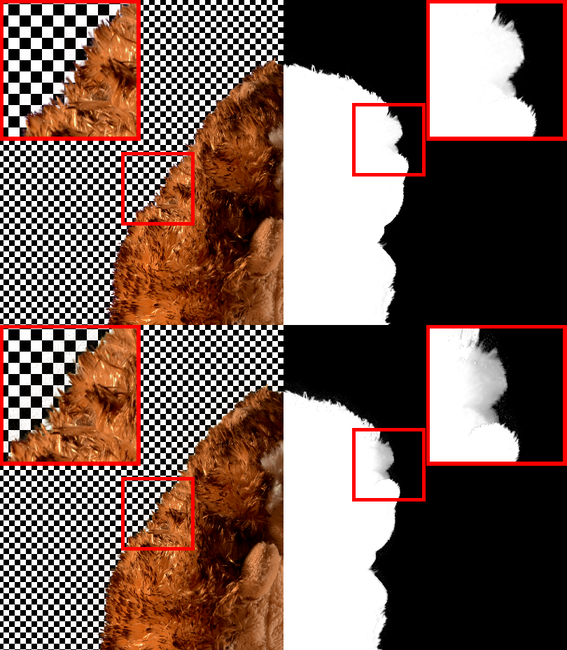}
    }	    
  \end{subfigure}
  \begin{subfigure}
    {
    \centering
    \includegraphics[height=\myheight\textheight]{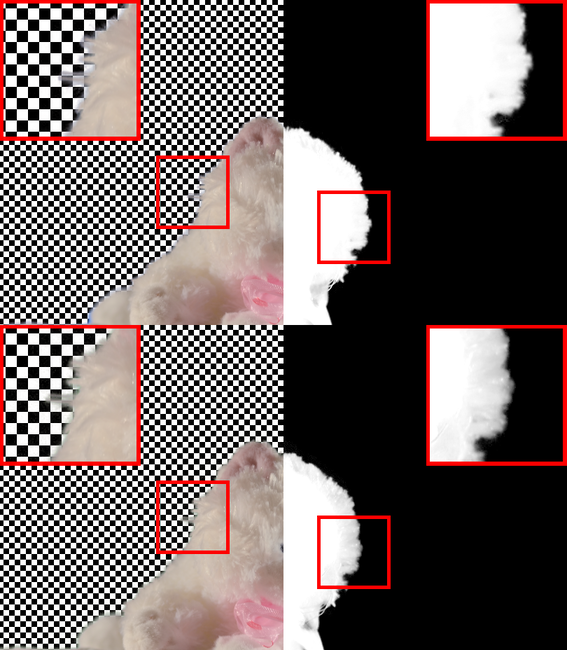}
    }
  \end{subfigure}
  \begin{subfigure}
    {
    \centering
    \includegraphics[height=\myheight\textheight]{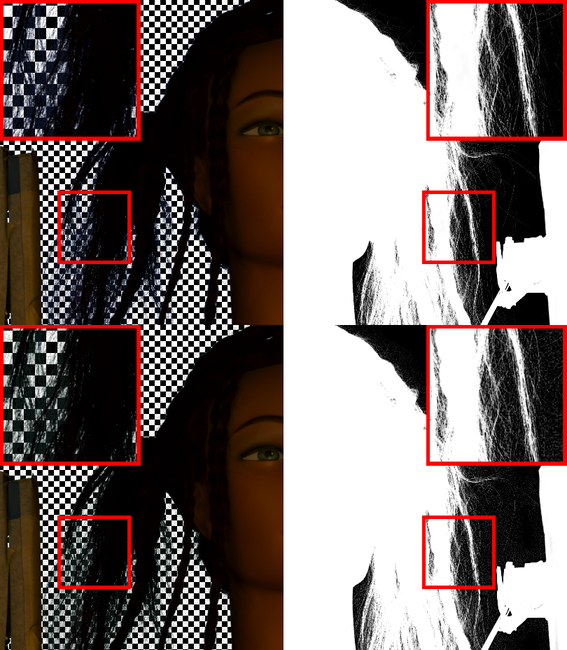}
    }
  \end{subfigure}
  \begin{subfigure}
    {
    \centering
    \includegraphics[height=\myheight\textheight]{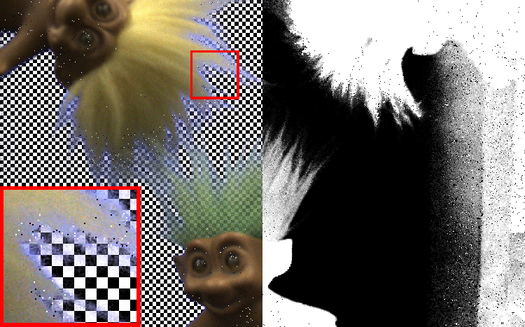}
    }
  \end{subfigure}  
  \hspace{-4mm}
  \begin{subfigure}
    {
    \centering
    \includegraphics[height=\myheight\textheight]{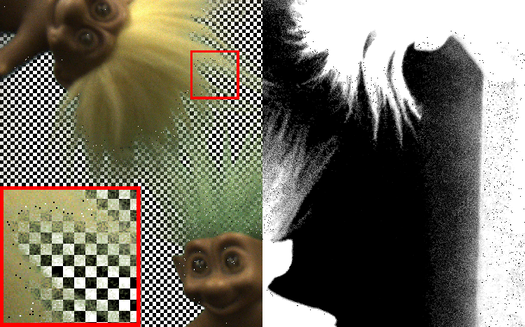}
    }
  \end{subfigure}
  \\[-2mm]
  \rotatebox{90}{\centering{\color{white}Xg...\color{black}SIM \cite{Sun:2021}}}
  \begin{subfigure}
    {
    \centering
    \includegraphics[height=\myheight\textheight]{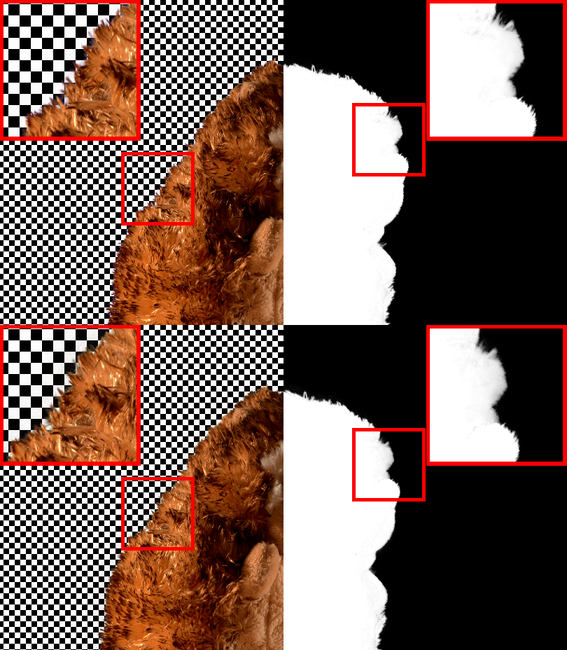}
    }	    
  \end{subfigure}
  \begin{subfigure}
    {
    \centering
    \includegraphics[height=\myheight\textheight]{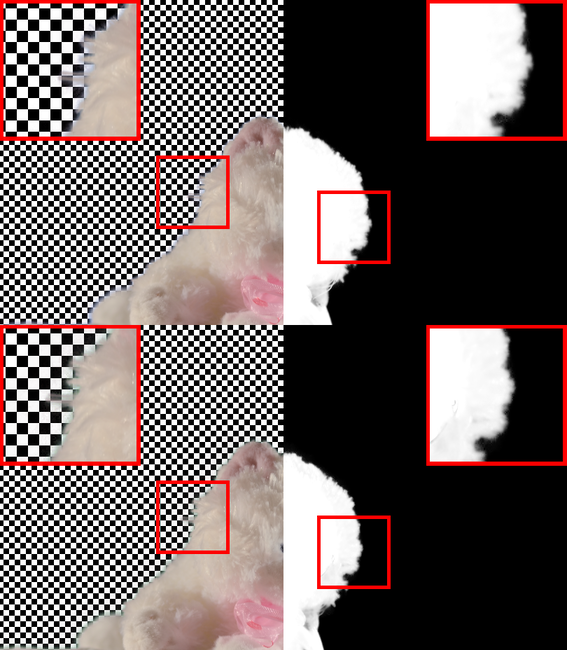}
    }
  \end{subfigure}
  \begin{subfigure}
    {
    \centering
    \includegraphics[height=\myheight\textheight]{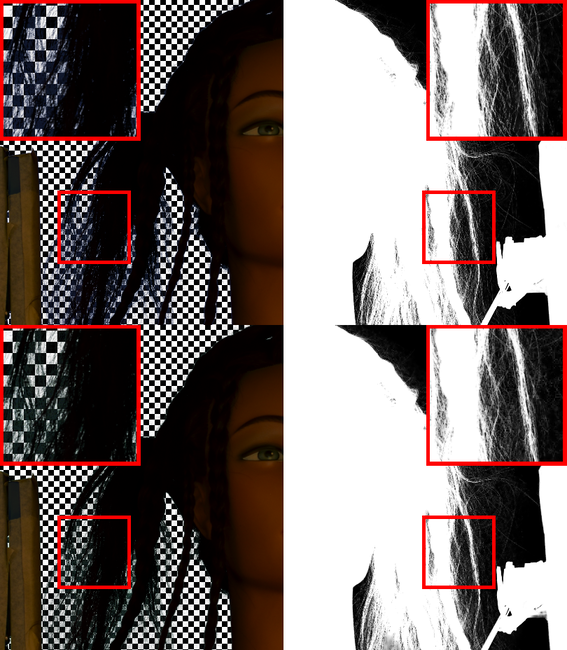}
    }
  \end{subfigure}
  \begin{subfigure}
    {
    \centering
    \includegraphics[height=\myheight\textheight]{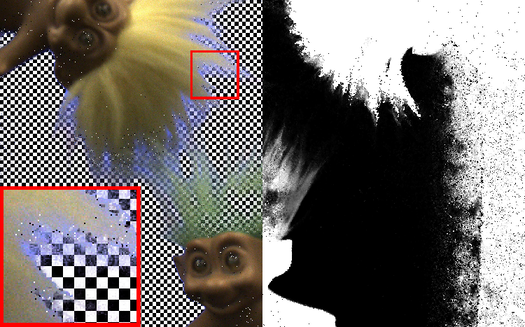}
    }
  \end{subfigure}  
  \hspace{-4mm}
  \begin{subfigure}
    {
    \centering
    \includegraphics[height=\myheight\textheight]{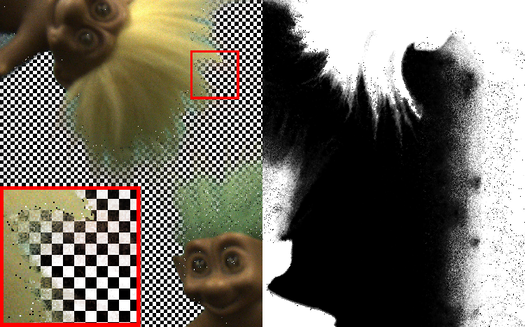}
    }
  \end{subfigure} 
  \\[-2mm]
  \rotatebox{90}{\centering{\color{white}Xg...\color{black}BSM$^6$ \cite{Smith:1996}}}  
  \begin{subfigure}
    {
    \centering
    \includegraphics[height=\myheight\textheight]{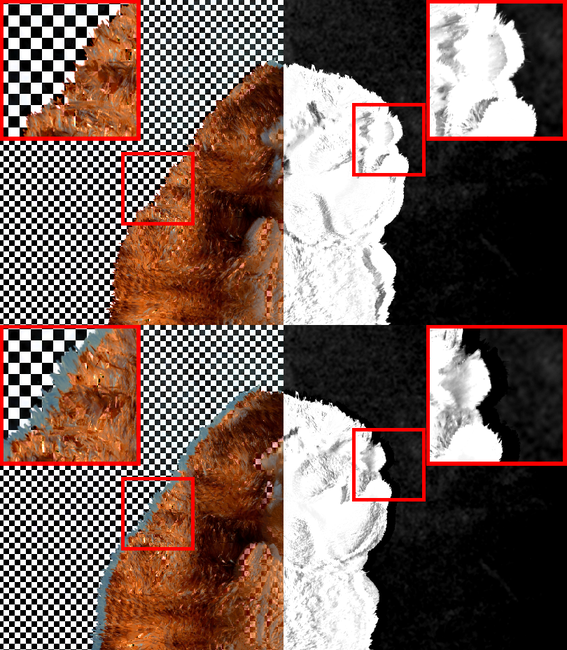}
    }
  \end{subfigure}
  \begin{subfigure}
    {
    \centering
    \includegraphics[height=\myheight\textheight]{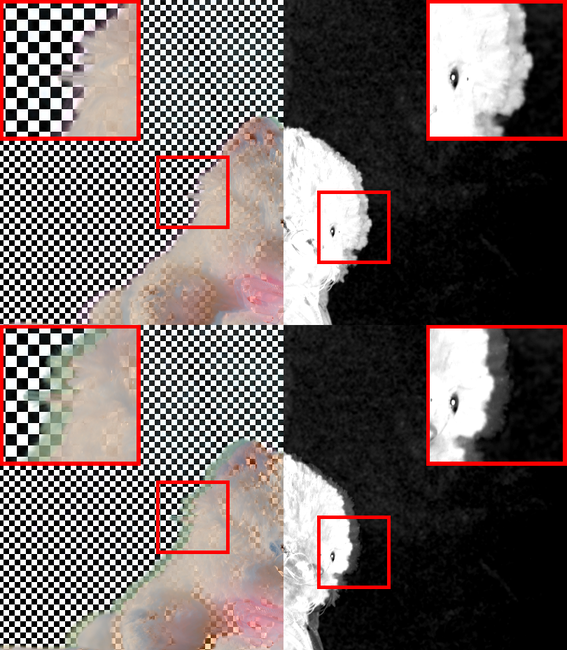}
    }	    
  \end{subfigure}
  \begin{subfigure}
    {
    \centering
    \includegraphics[height=\myheight\textheight]{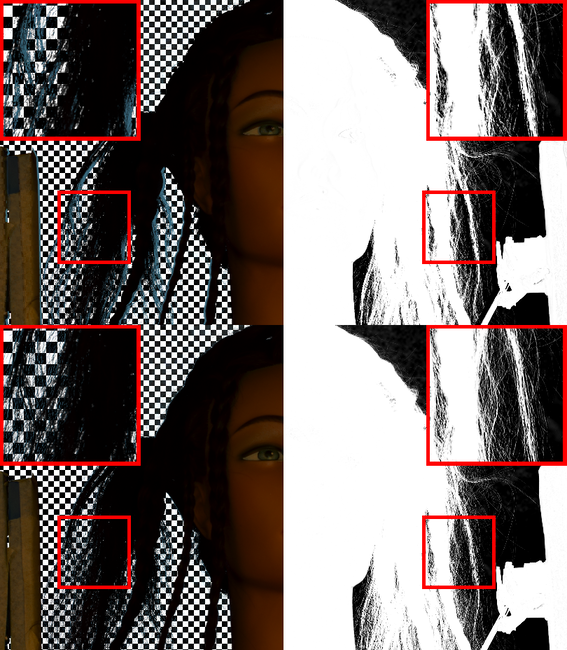}
    }
  \end{subfigure}
  \begin{subfigure}
    {
    \centering
    \includegraphics[height=\myheight\textheight]{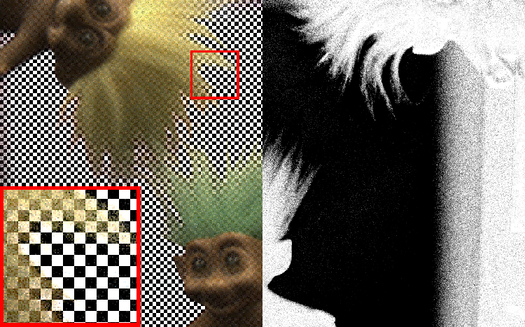}
    }
  \end{subfigure} 
  \hspace{-4mm}
  \begin{subfigure}
    {
    \centering
    \includegraphics[height=\myheight\textheight]{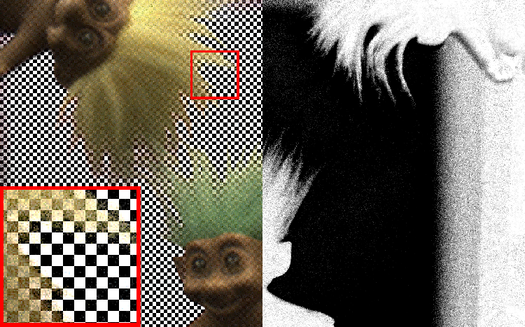}
    }
  \end{subfigure} 
  \\[-2mm]
  \rotatebox{90}{\centering{\color{white}Xg....\color{black}CIM$^6$ \cite{Grundhoefer:2010}}}
  \begin{subfigure}
    {
    \centering
    \includegraphics[height=\myheight\textheight]{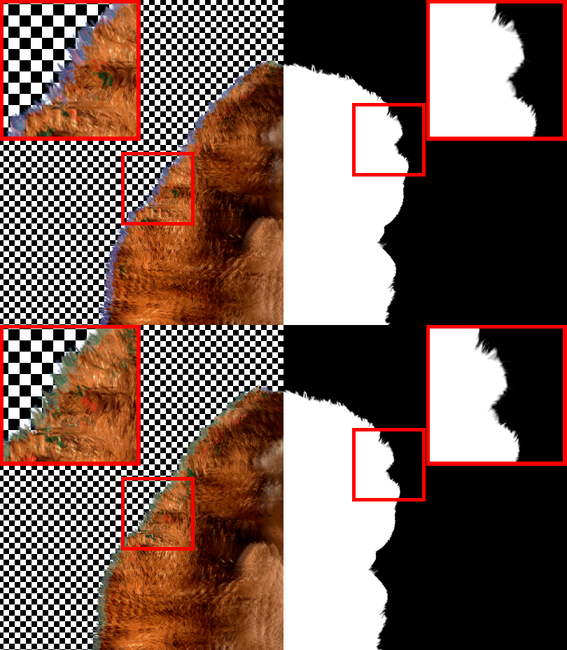}
    }
  \end{subfigure}
  \begin{subfigure}
    {
    \centering
    \includegraphics[height=\myheight\textheight]{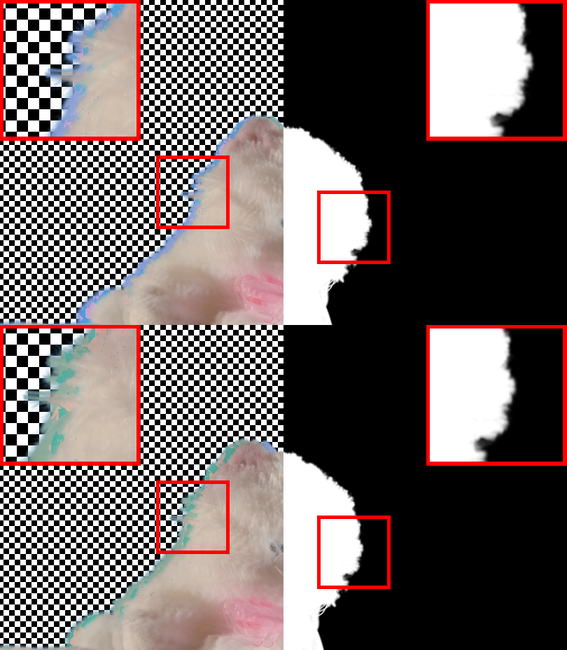}
    }	    
  \end{subfigure}
  \begin{subfigure}
    {
    \centering
    \includegraphics[height=\myheight\textheight]{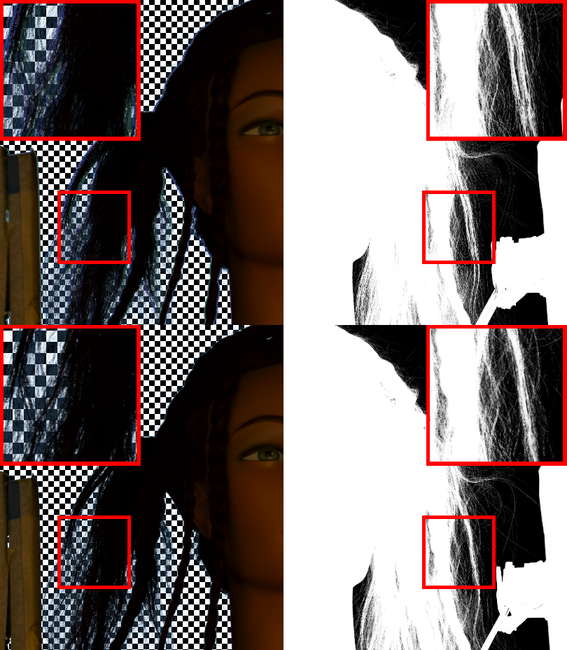}
    }
  \end{subfigure}
  \begin{subfigure}
    {
    \centering
    \includegraphics[height=\myheight\textheight]{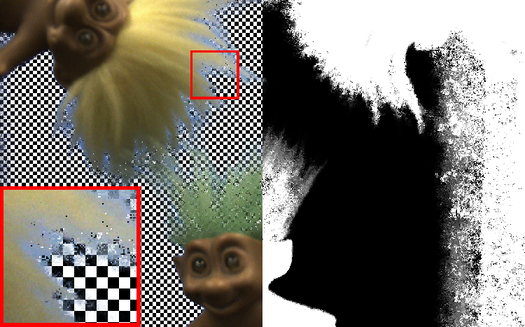}
    }
  \end{subfigure} 
  \hspace{-4mm}
  \begin{subfigure}
    {
    \centering
    \includegraphics[height=\myheight\textheight]{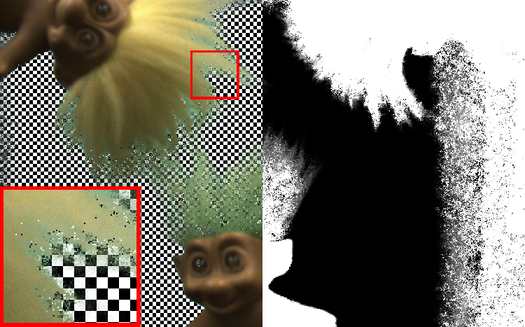}
    }
  \end{subfigure} 
  \\[-2mm]
  \rotatebox{90}{\centering{\color{white}Xg.......\color{black}ours$_{ma}$}}
  \begin{subfigure}[Dmitriy]
    {
    \addtocounter{subfigure}{-35}
    \centering
    \includegraphics[height=\myheight\textheight]{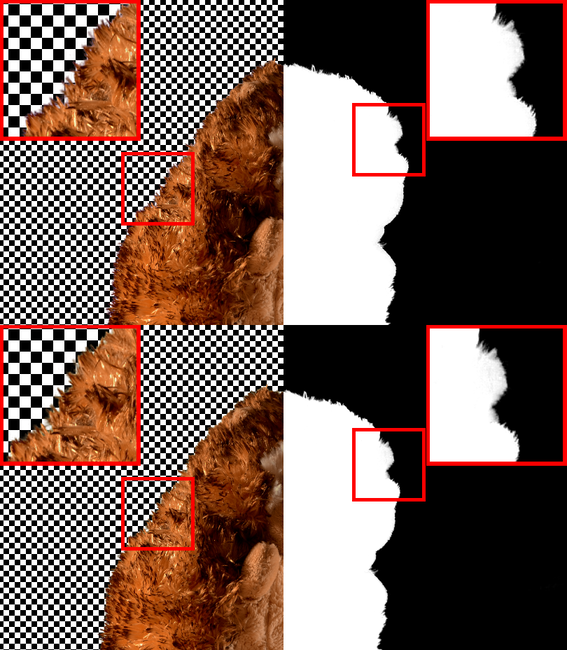}
    }	    
  \end{subfigure}
  \begin{subfigure}[Alex]
    {
    \centering
    \includegraphics[height=\myheight\textheight]{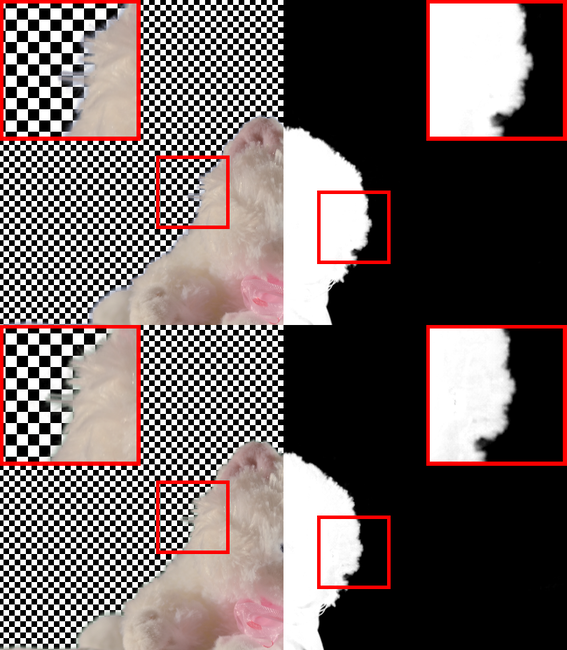}
    }
  \end{subfigure}
  \begin{subfigure}[Castle]
    {
    \centering
    \includegraphics[height=\myheight\textheight]{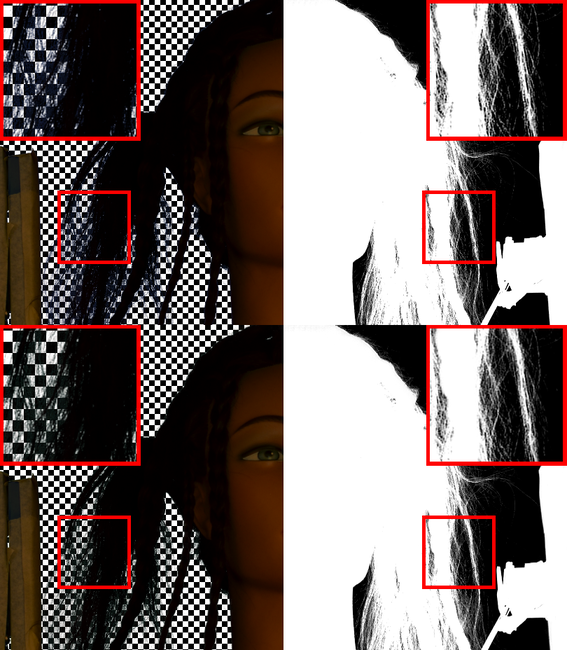}
    }
  \end{subfigure}
  \begin{subfigure}[Trolls with blue background]
    {
    \centering
    \includegraphics[height=\myheight\textheight]{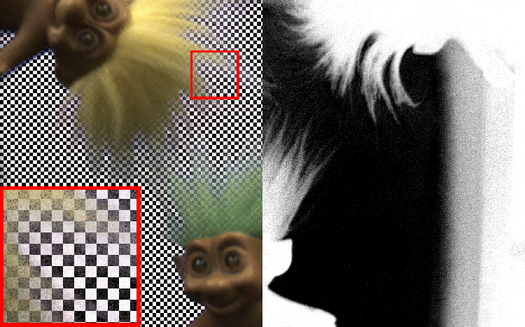}
    }
  \end{subfigure} 
  \hspace{-4mm}
  \begin{subfigure}[Trolls with green background]
    {
    \centering
    \includegraphics[height=\myheight\textheight]{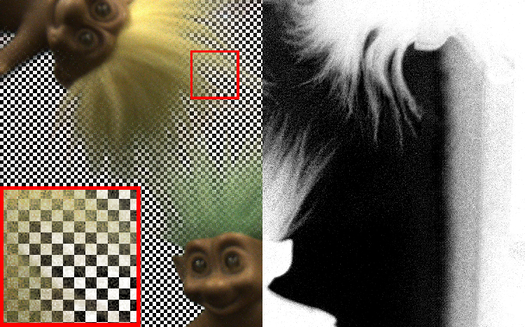}
    }
  \end{subfigure}  
  \vspace{-2mm}
  \caption{Qualitative comparison of five matting methods. Please zoom in for better comparison.}
  \label{fig:comparison}
  \end{center}
\end{figure*}
\clearpage

\bibliographystyle{ACM-Reference-Format}
\bibliography{mybibliography}

\end{document}